\documentclass[10pt,twocolumn,letterpaper]{article}

\usepackage{wacv}
\usepackage{times}
\usepackage{epsfig}
\usepackage{graphicx}
\usepackage{amsmath}
\usepackage{amssymb}
\usepackage{booktabs}
\usepackage[LGRgreek]{mathastext}  

\usepackage{caption}
\usepackage{subcaption}
\usepackage{dblfloatfix}
\def\wacvPaperID{****} 

\wacvfinalcopy 

\ifwacvfinal
\usepackage[breaklinks=true,bookmarks=false]{hyperref}
\else
\usepackage[pagebackref=true,breaklinks=true,colorlinks,bookmarks=false]{hyperref}
\fi

\begin{document}

\title{Hyperspectral Image Compression Using Implicit Neural Representation}

\author{Shima Rezasoltani\\
Ontario Tech University\\
\and
Faisal Z. Qureshi\\
Ontario Tech University\\
}

\maketitle
\thispagestyle{empty}

\begin{abstract}

Hyperspectral images, which record the electromagnetic spectrum for a pixel in the image of a scene, often store hundreds of channels per pixel and contain an order of magnitude more information than a typical similarly-sized color image.   Consequently, concomitant with the decreasing cost of capturing these images, there is a need to develop efficient techniques for storing, transmitting, and analyzing hyperspectral images.  This paper develops a method for hyperspectral image compression using implicit neural representations where a multilayer perceptron network $\Phi_\theta$ with sinusoidal activation functions ``learns'' to map pixel locations to pixel intensities for a given hyperspectral image $I$.  $\Phi_\theta$ thus acts as a compressed encoding of this image.  The original image is reconstructed by evaluating $\Phi_\theta$ at each pixel location.  We have evaluated our method on four benchmarks---Indian Pines, Cuprite, Pavia University, and Jasper Ridge---and we show the proposed method achieves better compression than JPEG, JPEG2000, and PCA-DCT at low bitrates.

\end{abstract}

\section{Introduction}
Hyperspectral images capture electromagnetic spectrum for a pixel in the image of a scene~ \cite{l1971lightness,goetz1985imaging}.  These images offer many possibilities for object detection, material identification, and scene analysis.  The costs associated with capturing high-resolution temporal, spatial, and spectral data continue to decrease, and it is no surprise that hyperspectral images have found widespread use in areas such as, remote sensing, biotechnology, crop analysis, environmental monitoring, food production, medical diagnosis, pharmaceutical industry, mining, and oil \& gas exploration, etc.~\cite{ghamisi2017advances,govender2007review,adam2010multispectral,fischer2006multispectral,liang2012advances,carrasco2003hyperspectral,afromowitz1988multispectral, kuula2012using,schuler2012preliminary,padoan2008quantitative,edelman2012hyperspectral,gowen2007hyperspectral,feng2012application,clark1995mapping}.  Unlike color images that record red, green, and blue channels per pixel, hyperspectral images record 100s of channels per pixel, representing pixels’ spectra.  This suggests that hyperspectral images require two orders of magnitude more space than what is needed to store similarly-sized color images.  Consequently, there is a need to develop efficient schemes for capturing, storing, transmitting, and analyzing hyperspectral images.  This work studies hyperspectral image compression which plays an important role in the storage and transmission of these images.

\par Recently, there has been a surge in interest in learning-based compression schemes.  For example, autoencoders~\cite{Hz94} and rate-distortion autoencoders~\cite{APF18, BLS17} have been used to learn compact representations of the input signals.  Here network weights together with the signal signature---latent representation in the case of autoencoders---serve as the compressed representation of the input signal.  Other concurrent works are exploring the use of Implicit Neural Representations (INRs) for signal compression.  INRs are well-suited to represent data that lives on an underlying grid, and these offer a new paradigm for signal representation.  The goal here is to learn a mapping between a location, say an $(x,y)$ pixel location for a 2D-image $I$, and the signal value at that location $I[x,y]$.  This mapping is subsequently used to recreate the original signal.  It is as simple as evaluating the INR at various locations.  In the case of INRs, network weights serve as the learned representation of the input signal.

\par We investigate the use of INRs for hyperspectral image compression and show that it is possible to achieve high rates of compression while maintaining acceptable reconstruction quality.  We evaluate the proposed approach on four benchmarks---Jasper Ridge, Indian Pines, Pavia University, and Cuprite---against three approaches---JPEG~\cite{good1994joint, qiao2014effective}, PCA-DCT~\cite{nian2016pairwise}, and JPEG2000~\cite{du2007hyperspectral} ---that others have used to compress hyperspectral data.  The results confirm that our method achieves a better Peak Signal-to-Noise ratio (PSNR) at low compression rates than those obtained by other methods.

\par The rest of the paper is organized as follows.  We discuss related works in the next section. Section Method describes the Image compression using implicit neural representation method and the pipeline of our proposed method. Next, we present the experimental setup and discuss the results. Section conclusions conclude a paper with a summary and possible directions for future work.


\begin{figure*}[t]
\begin{center} 
    \includegraphics[width=0.8\linewidth]
    {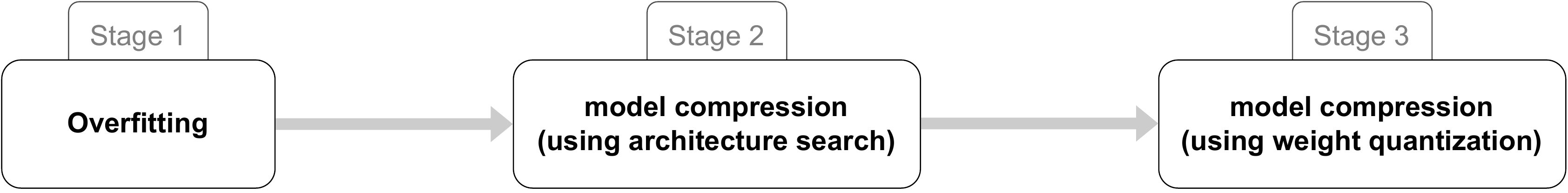}
\end{center}
   \caption{Overview of INR-based compression pipeline comprising overfitting, model compression using architecture search, and model compression using weight quantization.}
\label{fig:pip1}
\end{figure*}

\section{Related work}

Hyperspectral images exhibit both spatial and spectral redundancies that can be exploited to achieve compression.  Lossless compression schemes---e.g., those that use quantization or rely upon entropy-coding or statistics-based schemes---where it is possible to recover the original signal precisely often do not yield large savings~\cite{MH12,NV13}.  Lossy compression schemes, on the other hand, promise large savings while maintaining acceptable reconstruction quality.   Inter-band compression techniques aim to eliminate spectral redundancy~\cite{Cha13}, while intra-band compression techniques aim to exploit spatial correlations.  Intra-band compression techniques often follow the ideas developed for typical color image compression.  \cite{ZZW16} exploit the fact that groups of pixels that are around the same location in two adjacent bands are strongly correlated and proposes schemes that perform both inter-band and intra-band compression.  Principal Component Analysis (PCA) based techniques are frequently used for hyperspectral image compression.  PCA offers strong spectral decorrelation and is used to reduce the number of channels in a hyperspectral image.  The remaining channels are subsequently compressed using Joint Picture Expert Group (JPEG) or JPEG2000 standard~\cite{BS10, DF07, WWJ09, PTM07}.  Along similar lines, tensor decomposition methods have also been applied to the problem of hyperspectral image compression~\cite{ZZT15}.  Tensor decomposition achieves dimensionality reduction while maintaining the spatial structure.  Transform coding schemes that achieve image compression by reducing spatial correlation have also been used to compress hyperspectral data.  Discrete Cosine Transform (DCT) has been used to perform intra-band compression; however, it ignores iner-band (or spectral) redundancy.  3D-DCT that divides a 3-dimensional hyperspectral image into $8\times8\times8$ datacubes is proposed to achieve both inter-band and intra-band compression~\cite{QRS14}.  Similarly to JPEG, which uses $8\times8$ blocks, 3D-DCT exhibits blocking effects in reconstructed hyperspectral images.  The blocking effects can be removed to some extent by using wavelet transform instead~\cite{RSU12,GBS09}.  Video coding methods coupled with inter-band spectral prediction modeling have also been proposed to compress hyperspectral images.  

\section{Method}
\subsection{Background}
In the 3D vision field, interest in using neural networks to represent data has recently increased \cite{niemeyer2019occupancy, park2019deepsdf, chen2019learning}, after the original proposal made by \cite{stanley2007compositional}. High-resolution signals can be compactly encoded via implicit representations, as suggested by studies \cite{mildenhall2020nerf, sitzmann2020implicit, tancik2020fourier}.
\par Commonly, learned image compression techniques use hierarchical variational autoencoders \cite{balle2018variational, lee2018context, minnen2018joint}, where the latent variables are discretized for entropy coding purpose. Paper \cite{dupont2021coin} proposes a different approach for RGB image compression. Their method stores the weights of a neural network overfitted to the image.  Similar to previous research on latent variable models \cite{hjelm2016iterative, kim2018semi, krishnan2018challenges, marino2018iterative}, numerous studies \cite{campos2019content, guo2020variable, yang2020improving} make an effort to close the amortization gap \cite{cremer2018inference} by combining the usage of amortized inference networks with iterative gradient-based optimization procedures. Using inference time per instance optimization, \cite{yang2020improving} also identifies and makes an effort to bridge the discretization gap caused by quantizing the latent variables. The concept of per-instance model optimization is expanded upon in research \cite{van2021overfitting}, which fine-tunes the decoder for each instance and transmits updates to the quantized decoder's parameters together with the latent code to provide better rate-distortion performance.

\begin{figure*}[t]
\begin{center} 
    \includegraphics[width=0.8\linewidth]{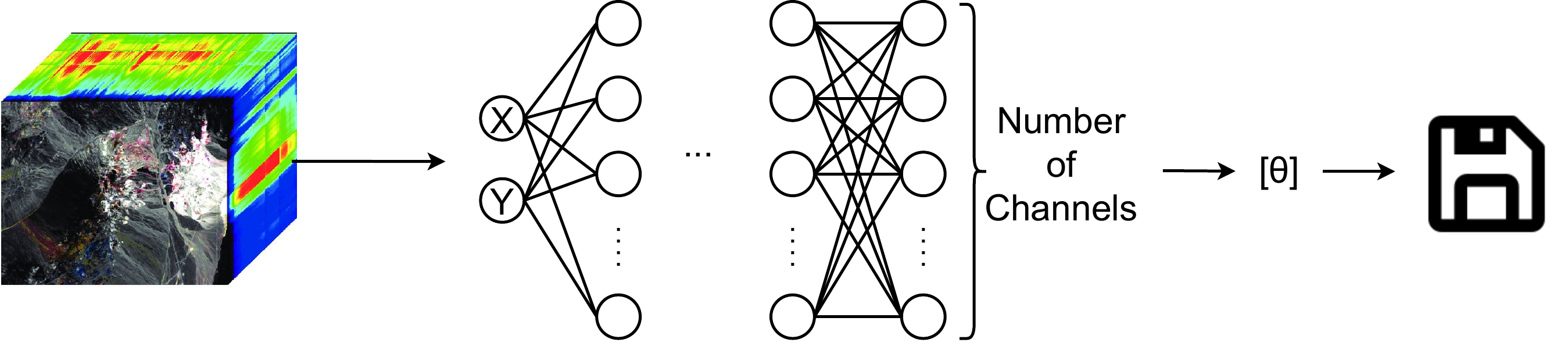}
\end{center}
   \caption{Compressed neuronal representations that are implicit. A neural network that maps pixel positions (x, y) to channel values are used to overfit an image. The weights of this neural network are then transmitted after being quantized to a smaller bit-width.}
\label{fig:n1}
\end{figure*}

\subsection{Image Compression using INRs}
INRs represent data as a continuous function from coordinates to values, storing coordinate-based data like photos, videos, and 3D forms. For instance, hyperspectral image maps to a channel vector in a channel space based on the horizontal and vertical coordinates $(p_x, p_y)$:
\begin{equation}
I : (p_x, p_y) \rightarrow (ch_0, ch_1, ..., ch_n),
\end{equation}
In which ch refers to channels, and n equals the number of channels for each image.
A neural network $f_\Theta$, often a Multi-Layer Perceptron (MLP) with parameters $\Theta$, can be used to approximate this mapping such that $I(p_x, p_y) \approx f_\Theta(p_x, p_y)$.
INRs can be evaluated on any coordinates inside the normalized range [1, 1] since they are continuous functions.
\par INRs implicitly store all data in the network weights. The coordinate, which is used as the input to the INR, contains no information. The INR is trained through the encoding procedure. Decoding is the same as adding a set of weights to the network and assessing the results on a coordinate grid. This can be summed up as:
\begin{equation}
    \arg\min_\theta L(x,f_\theta(p)) = \theta^* \xrightarrow[transmit \theta^*]{} \widehat{X} = f_{\theta^*}(p).
\end{equation}
In order to recreate a distorted replica of the original image X, we just need to store $\Theta^\star$. With our strategy, we are able to simultaneously accomplish compact storage and effective repair.

\begin{figure}[b]
\begin{center} 
    \includegraphics[width=0.35\linewidth]
    {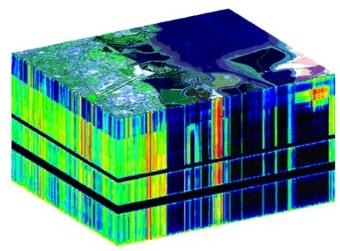}
\end{center}
   \caption{The JPL’s AVIRIS hyperspectral data cube provided on a NASA ER-2 plane over Moffett Field}
\label{fig:d1}
\end{figure}

\begin{figure*}[!ht]
     \centering
     \begin{subfigure}[b]{0.23\textwidth}
         \centering
         \includegraphics[width=\textwidth]{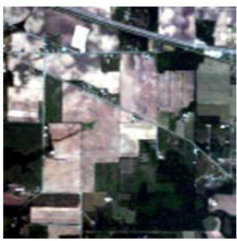}
         \caption{}
         \label{fig:d2}
     \end{subfigure}
     \hfill
     \begin{subfigure}[b]{0.23\textwidth}
         \centering
         \includegraphics[width=\textwidth]{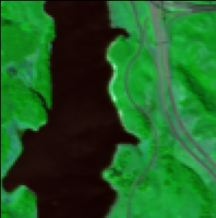}
         \caption{}
         \label{fig:d3}
     \end{subfigure}
     \hfill
     \begin{subfigure}[b]{0.16\textwidth}
         \centering
         \includegraphics[width=\textwidth]{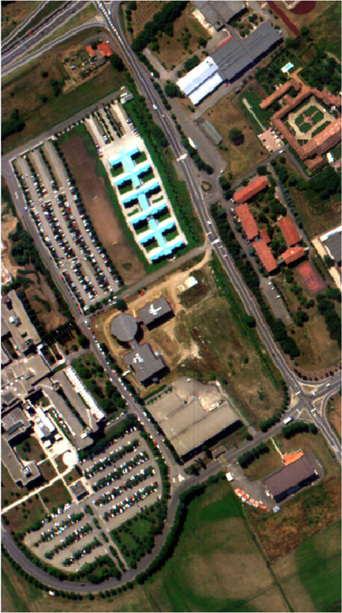}
         \caption{}
         \label{fig:d4}
     \end{subfigure}
     \hfill
     \begin{subfigure}[b]{0.28\textwidth}
         \centering
         \includegraphics[width=\textwidth]{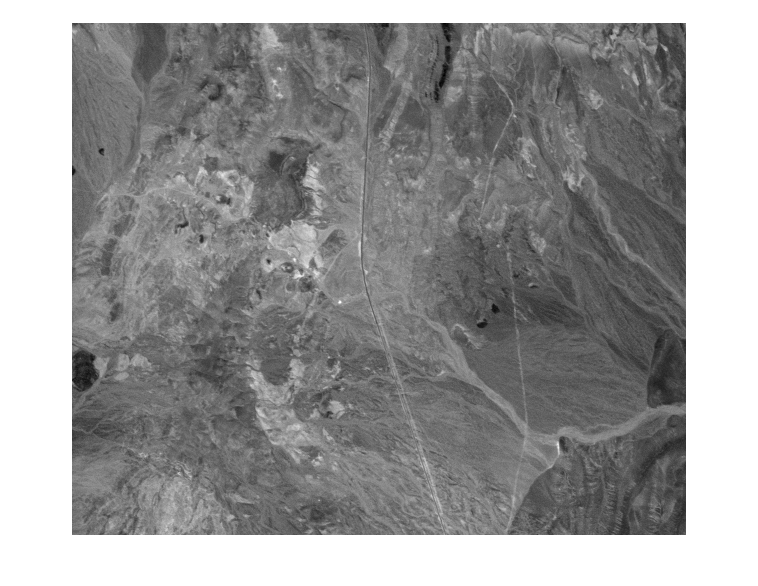}
         \caption{}
         \label{fig:d5}
     \end{subfigure}
     \caption{From left to right, datasets: Indian Pines, Jasper Ridge, Pavia University, and Cuprite.}
\end{figure*}

\subsection{Compression Pipeline for INRs}
The pipeline for our INR-based compression is described in this section. Figure \ref{fig:pip1} represent the entire pipeline and figure \ref{fig:n1} shows a higher-level overview.
\\\textbf{Stage 1: Overfitting.} We train an MLP to map pixel locations to pixel values in this work. In fact, at test time, we overfit the INR f to a data sample. This step is similar to calling the encoder in other learned techniques. To underline that the INR is trained only to represent one image, we refer to this stage as overfitting.

We save the weights of a neural network overfitted to the image rather than the pixel values for each pixel of a hyperspectral image. A general illustration of this work is proposed in Figure \ref{fig:n1}. To encode an image, we apply an MLP, which maps pixel locations to pixel values. We test the MLP at each pixel position to decode the image. While the extremely frequent data included in some hyperspectral images makes overfitting such multi-layer perceptrons, also known as implicit neural representations, challenging \cite{ronen2019convergence, tancik2020fourier}, recent studies have demonstrated that this can be minimized by employing sinusoidal encodings and activations \cite{tancik2020fourier, mildenhall2020nerf, sitzmann2020implicit}. As a result, in this study, we use MLPs with sine activations, often known as SIRENs \cite{sitzmann2020implicit}.
\\As a result, we first overfit an MLP to the hyperspectral image, then quantize and transmit its weights. To reconstruct the hyperspectral image, the transmitted MLP is then evaluated at all pixel positions.
Let I represent the hyperspectral image we want to encode, and $I[x, y]$ gives the pixel values at that specific pixel position for the encoding part $(x, y)$. In the hyperspectral image, we build a function $f_\Theta$ with parameters $\Theta$ mapping pixel positions to pixel band values. The hyperspectral image can then be encoded by overfitting $f_\Theta$ to it under some distortion control.
Given an image x and a coordinate grid p, we minimize the objective:
\begin{equation}
    \arg\min_\theta L_{MSE}(x,f_\theta(p)).
\end{equation}
The parameterization of $f_\Theta$ must be carefully chosen. Even when utilizing quite a few parameters, parameterizing $f_\Theta$ using an MLP with normal activation functions leads to underfitting \cite{tancik2020fourier, sitzmann2020implicit}. We choose the sine activation function for this work, according to \cite{sitzmann2020implicit}.
\\\textbf{Stage 2: Model Compression using Architecture Search.} 
 We run a hyperparameter tunning over the number of layers and layer size of the MLP and also quantize the weights from 32-bit to 16-bit precision to decrease the model size. As a result, we explore architecture search and weight quantization strategies to decrease model size.
\\\textbf{Stage 3: Model Compression using Weight Quantization.} Because the MLP parameters are stored as a compressed representation of the image, limiting the number of weights improves compression. The purpose is to fit $f_\Theta$ to I with the fewest parameters possible. The hyperspectral images are then compressed using model compression.
\\Decoding part is to just evaluate the function $f_\Theta$ at each pixel position to rebuild the image given the stored quantized weights. This decoding method provides several advantages, such as increased flexibility: we may decode the image in stages, for instance, by decoding portions of it or a low-resolution image initially, just by assessing $f_\Theta$ at different pixel positions. With autoencoder-based approaches, partly decoding images in this manner is challenging, demonstrating yet another benefit of this method.

\begin{figure*}
     \centering
     \begin{subfigure}[b]{0.46\textwidth}
         \centering
         \includegraphics[width=\textwidth]{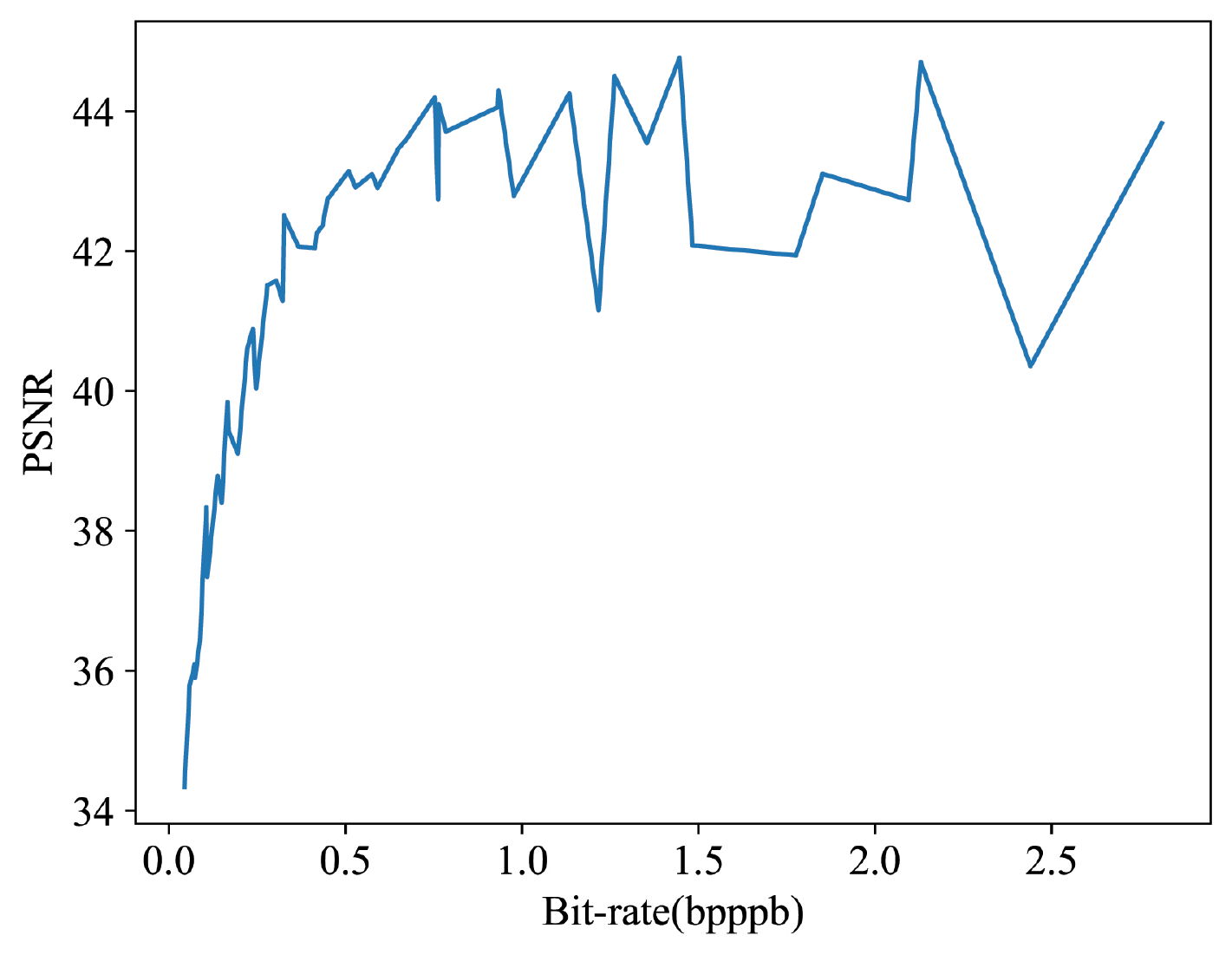}
         \caption{Indian Pines dataset.}
         \label{fig:BP1}
     \end{subfigure}
     \hfill
     \begin{subfigure}[b]{0.46\textwidth}
         \centering
         \includegraphics[width=\textwidth]{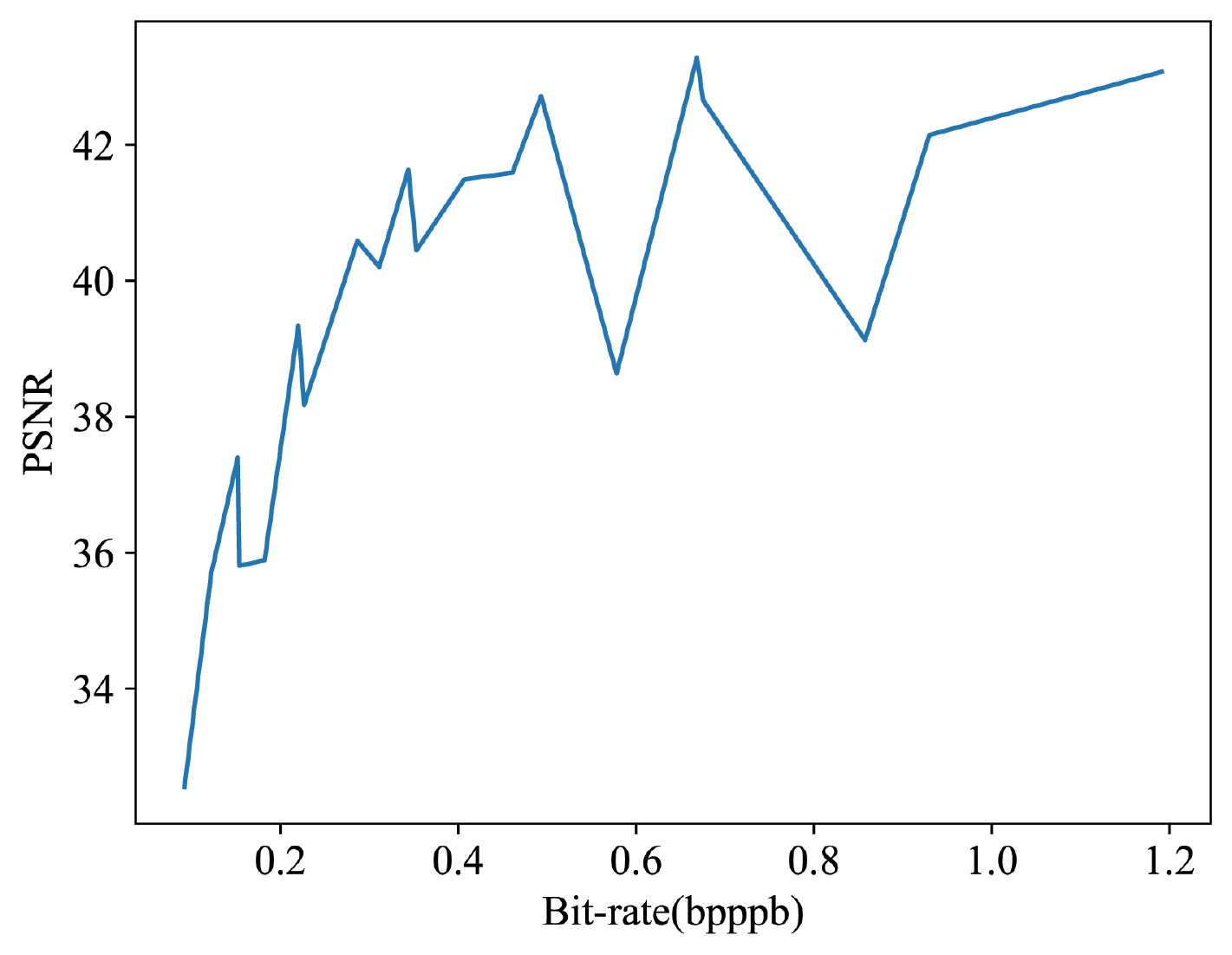}
         \caption{Jasper Ridge dataset.}
         \label{fig:BP2}
     \end{subfigure}
     \hfill
     \begin{subfigure}[b]{0.46\textwidth}
         \centering
         \includegraphics[width=\textwidth]{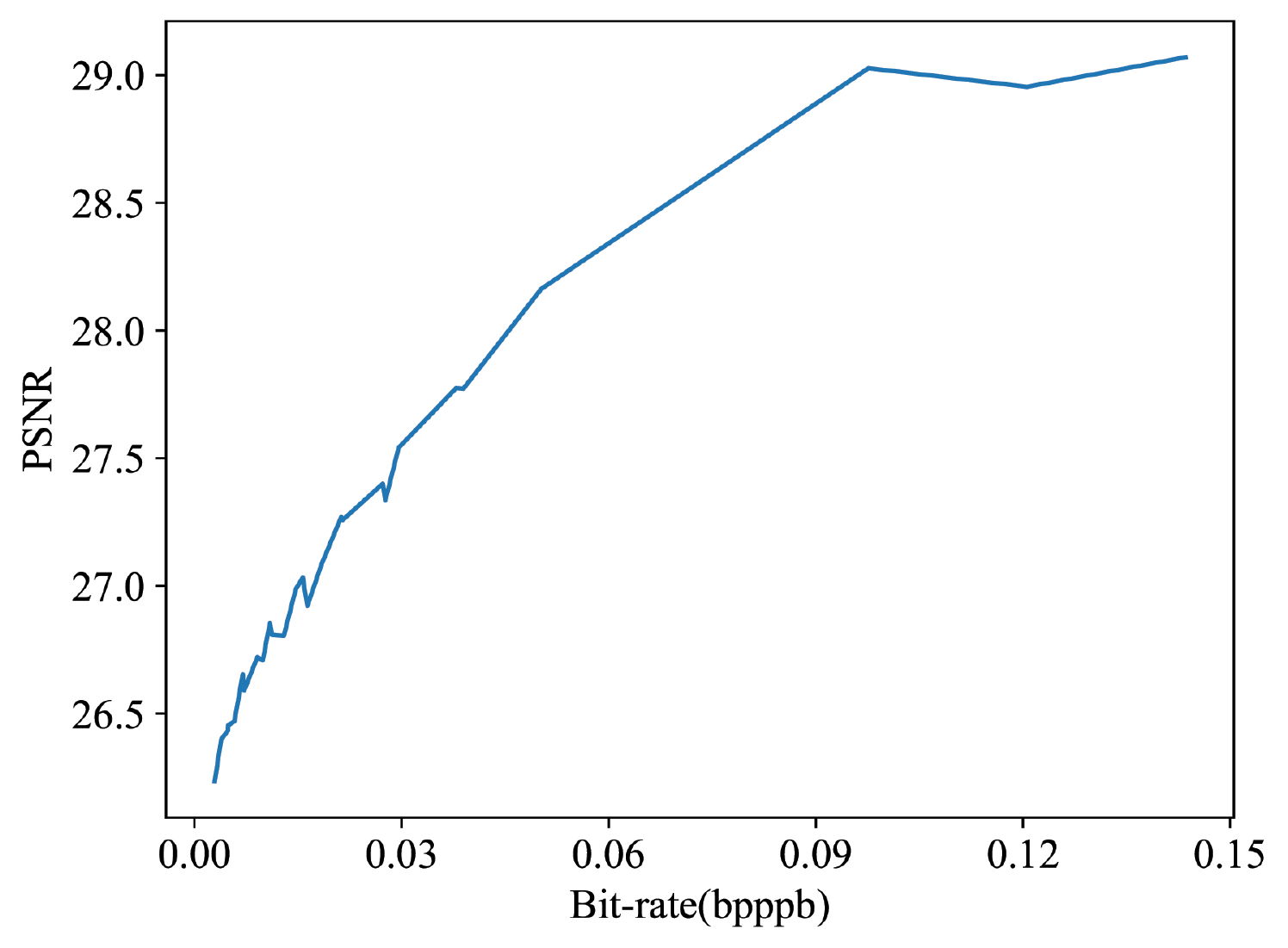}
         \caption{Cuprite dataset.}
         \label{fig:BP3}
     \end{subfigure}
     \hfill
     \begin{subfigure}[b]{0.46\textwidth}
         \centering
         \includegraphics[width=\textwidth]{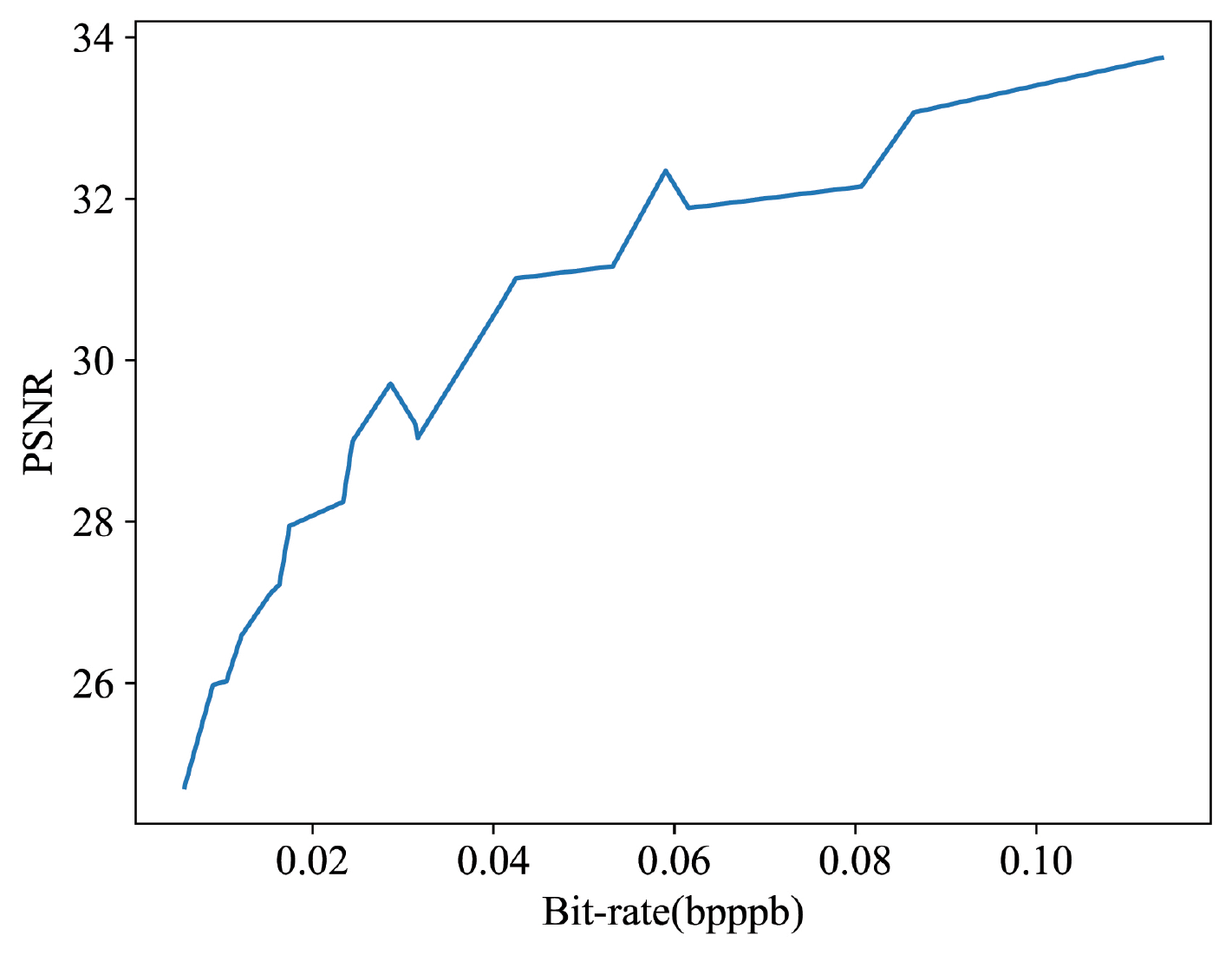}
         \caption{Pavia University dataset.}
         \label{fig:BP4}
     \end{subfigure}
     \caption{Rate distortion plots on different datasets.}
\end{figure*}

\subsection{Datasets}
We perform experiments on Indian Pines, Cuprite, Jasper Ridge, and Pavia University datasets. The AVIRIS instrument gathered the Indian Pines, Cuprite, and Jasper Ridge datasets.
With the spatial dimensions in the face and the spectral dimension in the depth, respectively, the hyperspectral imaging sensors give a three-dimensional data structure known as a data cube. Figure \ref{fig:d1}  is a cube image taken by the AVIRIS satellite of the Jet Propulsion Laboratory (JPL) above Moffett Field in California. Figure \ref{fig:d1}'s false-color image on top of the cube depicts a complex structure in the water and evaporation ponds to the right. On the cube's top, you can also see the Moffett Field airport.

The AVIRIS sensor gathers information from geometrically coherent spectroradiometric data that can be used to characterize the Earth's surface and atmosphere. The studies of oceanography, environmental science, snow hydrology, geology, volcanology, soil and land management, atmospheric and aerosol studies, agriculture, and limnology can all benefit from the use of these data. Applications for assessing and monitoring environmental risks such as toxic waste, oil spills, and air, land, and water pollution are currently being developed. The observations can be transformed to ground reflectance data, which can subsequently be utilized for quantitative assessment of surface features, with the required calibration and adjustment for atmospheric factors.

\subsubsection{Indian Pines}
The Indian Pines dataset, which has 145 by 145 pixels and 16 ground-truth classes, was collected by the AVIRIS instrument in 1992. It was obtained in NW Indiana over a mixed agricultural and wooded area. There are 220 channels in this image. The Indian Pines setting is made up primarily of agriculture, with the remaining third being either forest or other types of perennial forest vegetation. Along with some low-density homes, other built objects, and minor roads, there are two significant dual-lane motorways, a rail line, and other built objects.
Figure \ref{fig:d2} shows an image of this dataset.


\subsubsection{Jasper Ridge}
One well-known hyperspectral image is Jasper Ridge. The Jet Propulsion Laboratory (JPL) captured it using the AVIRIS (Airborne Visible/Infrared Imaging Spectrometer) sensor. Its dimensions are 512 by 614 pixels. A total of 224 electromagnetic bands between 380 nm and 2,500 nm are recorded for each pixel.
Figure \ref{fig:d3} shows an image of this dataset.



\subsubsection{Pavia University}
The Pavia University dataset was captured by the ROSIS-03 aerial instrument above the Pavia University in Italy. The flight above Pavia, Italy, was conducted by the German Aerospace Centre (the German Aerospace Agency) as part of the HySens project, which is run and funded by the European Union. For Pavia Centre and Pavia University, there are 102 and 103 spectral bands, respectively. Pavia Centre is 1096 by 1096 pixel image, whereas Pavia University is 610 by 610 pixels. The Pavia University data set includes several classes, such as \emph{trees, asphalt, bitumen, gravel, metal sheet, shadow, bricks, meadow,} and \emph{dirt}, and it belongs to the Engineering School at the University of Pavia.
Figure \ref{fig:d4} shows an image of this dataset.
\subsubsection{Cuprite}
The Cuprite dataset covers the Cuprite in Las Vegas, Nevada, United States. There are 224 channels with wavelengths between 370 and 2480 nm. Figure \ref{fig:d5} shows an image of this dataset.

\begin{figure*}
     \centering
     \begin{subfigure}[b]{0.46\textwidth}
         \centering
         \includegraphics[width=\textwidth]{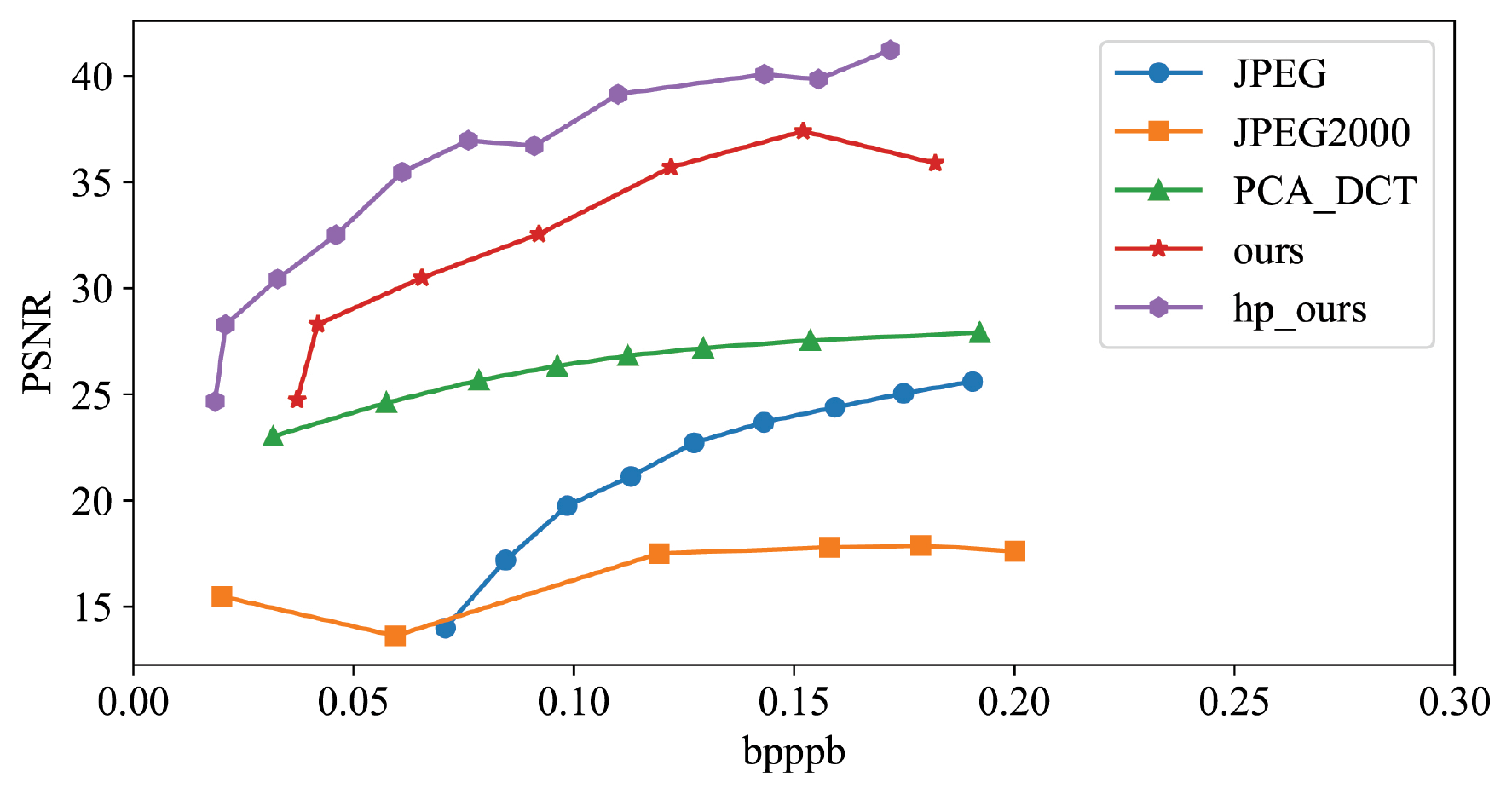}
         \caption{Jasper Ridge compression comparison plot. Our methods are compared with JPEG, JPEG2000, and PCA-DCT and outperform those benchmarks.}
         \label{fig:c1}
     \end{subfigure}
     \hfill
     \begin{subfigure}[b]{0.46\textwidth}
         \centering
         \includegraphics[width=\textwidth]{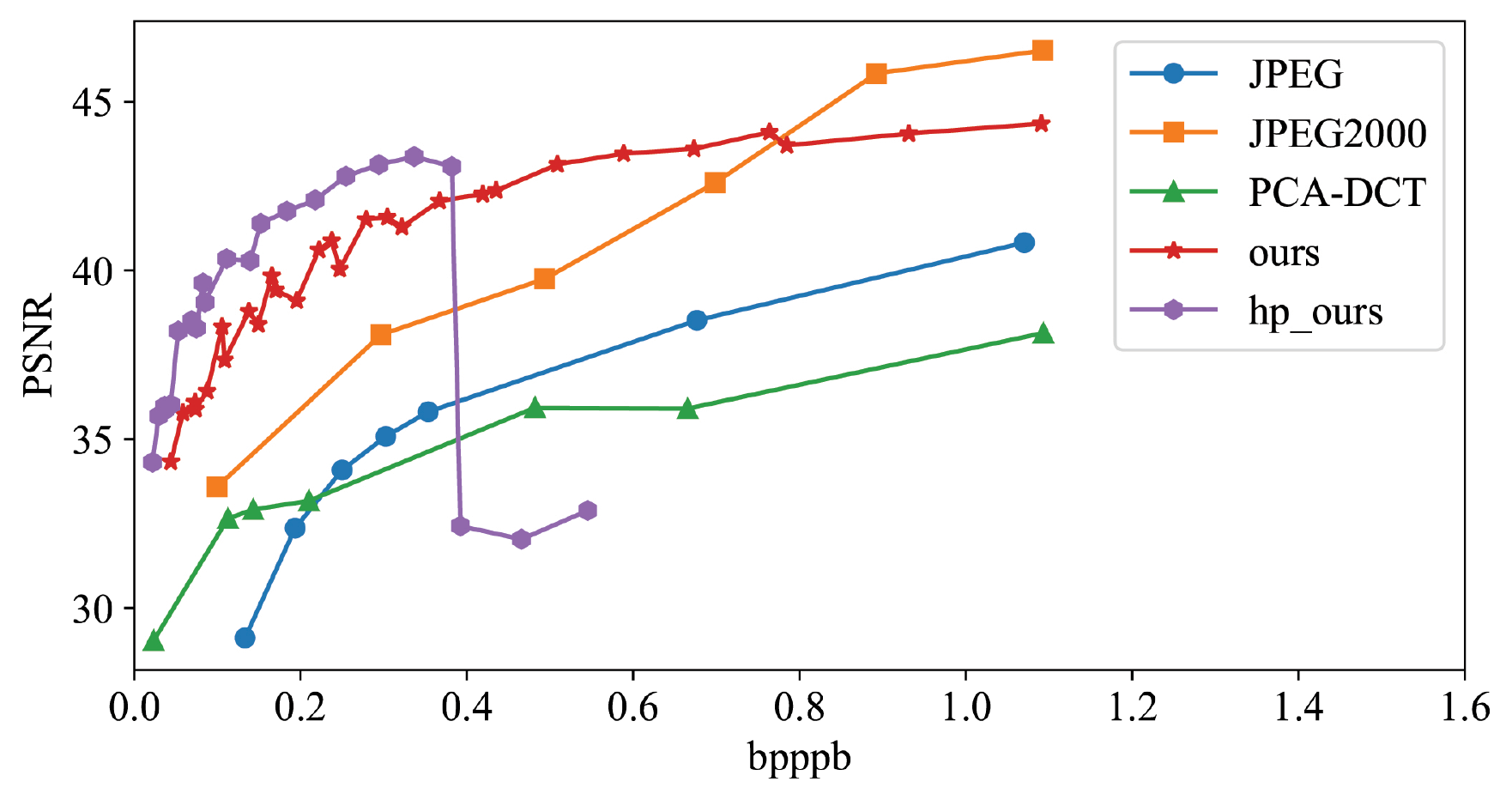}
         \caption{Indian Pines compression comparison plot. Our methods are compared with JPEG, JPEG2000, and PCA-DCT and outperform those benchmarks in full precision.}
         \label{fig:c2}
     \end{subfigure}
     \hfill
     \begin{subfigure}[b]{0.46\textwidth}
         \centering
         \includegraphics[width=\textwidth]{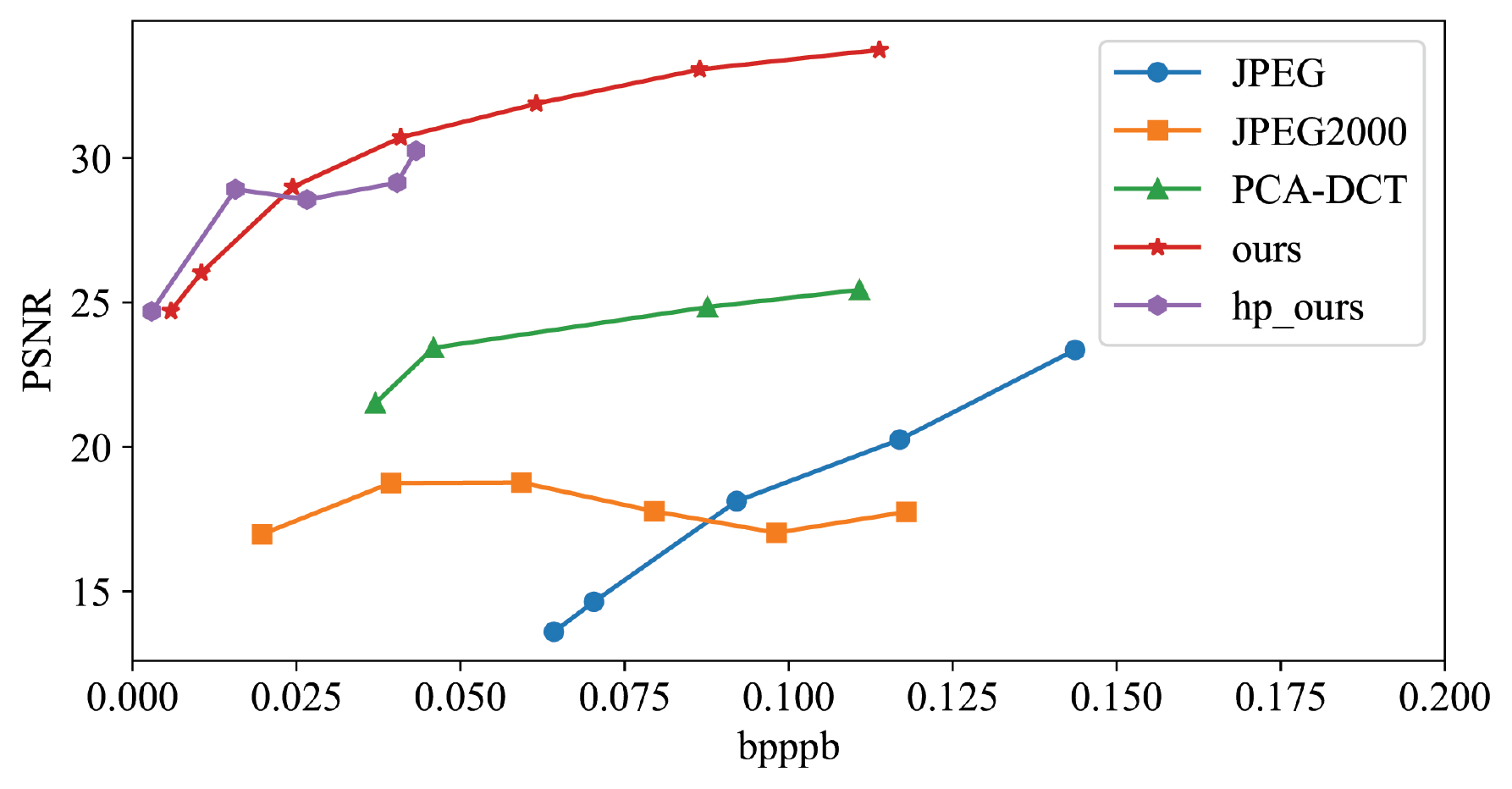}
         \caption{Pavia University compression comparison plot. Our methods are compared with JPEG, JPEG2000, and PCA-DCT and outperform those benchmarks.}
         \label{fig:c3}
     \end{subfigure}
     \hfill
     \begin{subfigure}[b]{0.46\textwidth}
         \centering
         \includegraphics[width=\textwidth]{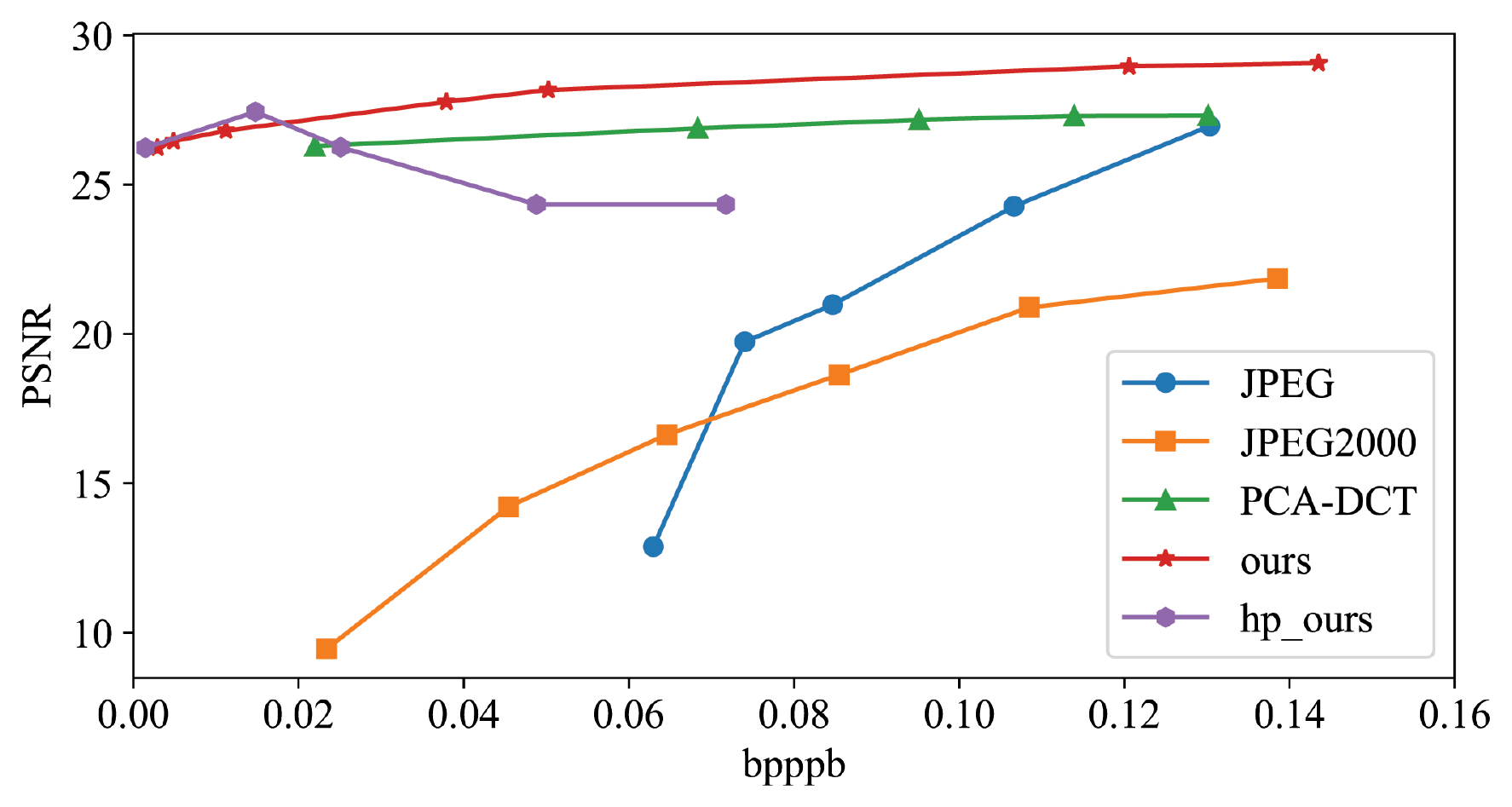}
         \caption{Cuprite compression comparison plot. Our methods are compared with JPEG, JPEG2000, and PCA-DCT and outperform JPEG and JPEG2000 at every bit-rates and PCA-DCT at low bitrates.}
         \label{fig:c4}
     \end{subfigure}
     \caption{Compression comparison plot for different datasets.}
\end{figure*}

\begin{figure*}
     \centering
     \begin{subfigure}[b]{0.46\textwidth}
         \centering
         \includegraphics[width=\textwidth]{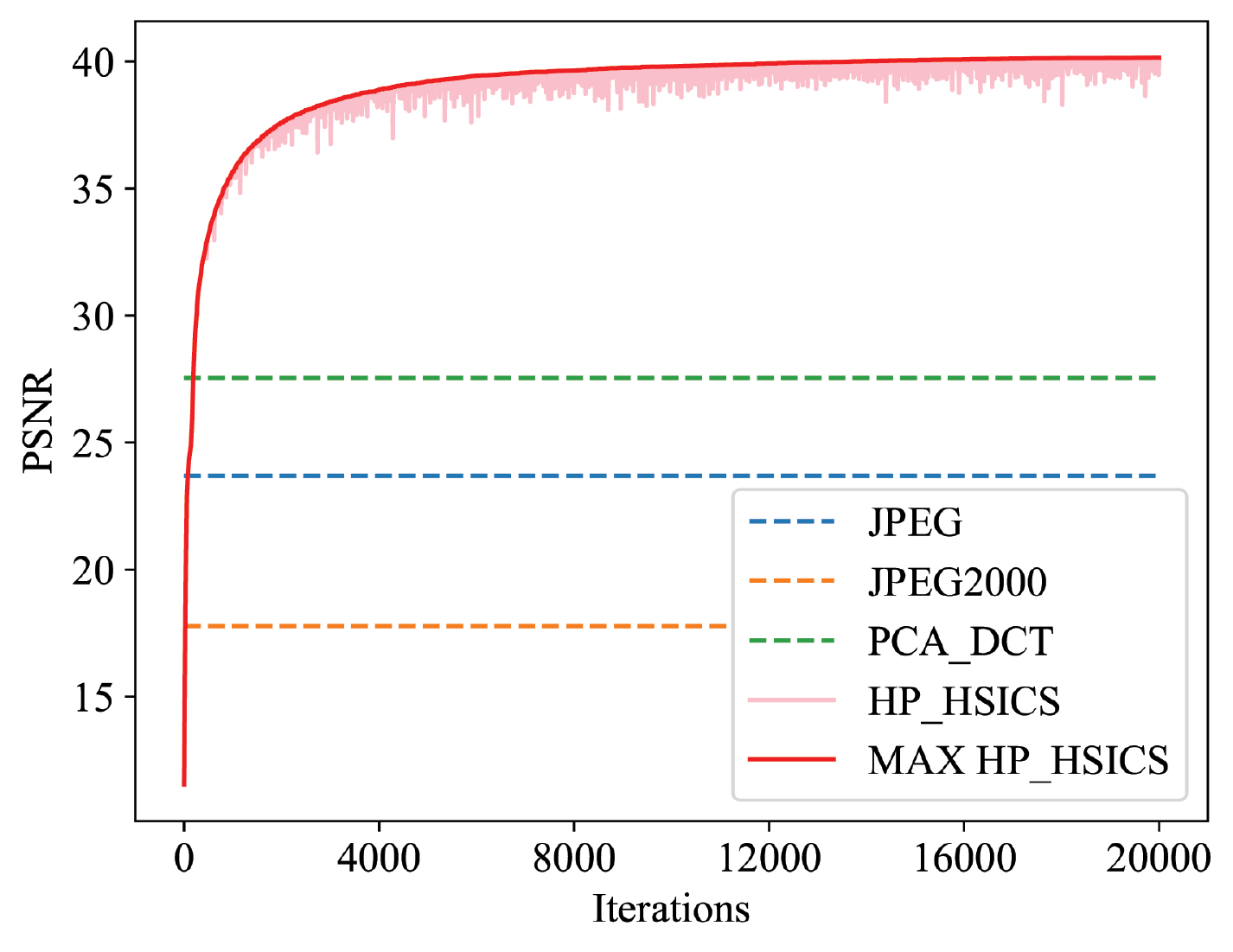}
         \caption{Model training on Jasper Ridge dataset at 0.15 bpppb. Our method outperforms JPEG, JPEG2000, and PCA-DCT methods after some iterations and continues improving beyond that.}
         \label{fig:ip1}
     \end{subfigure}
     \hfill
     \begin{subfigure}[b]{0.46\textwidth}
         \centering
         \includegraphics[width=\textwidth]{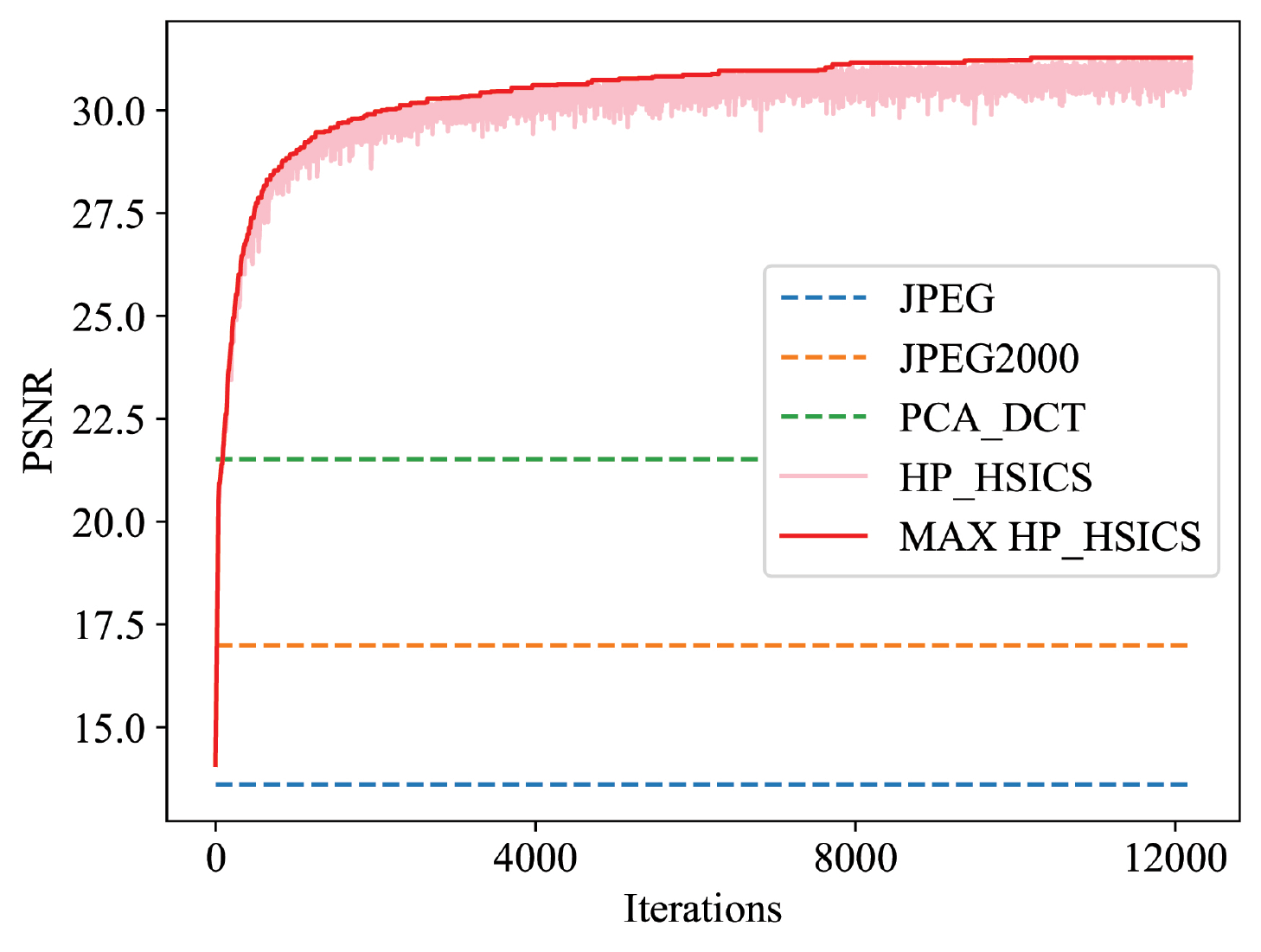}
         \caption{Model training on Pavia University dataset at 0.025 bpppb. Our method outperforms JPEG, JPEG2000, and PCA-DCT methods after some iterations and continues improving beyond that.}
         \label{fig:ip2}
     \end{subfigure}
     \hfill
     \begin{subfigure}[b]{0.46\textwidth}
         \centering
         \includegraphics[width=\textwidth]{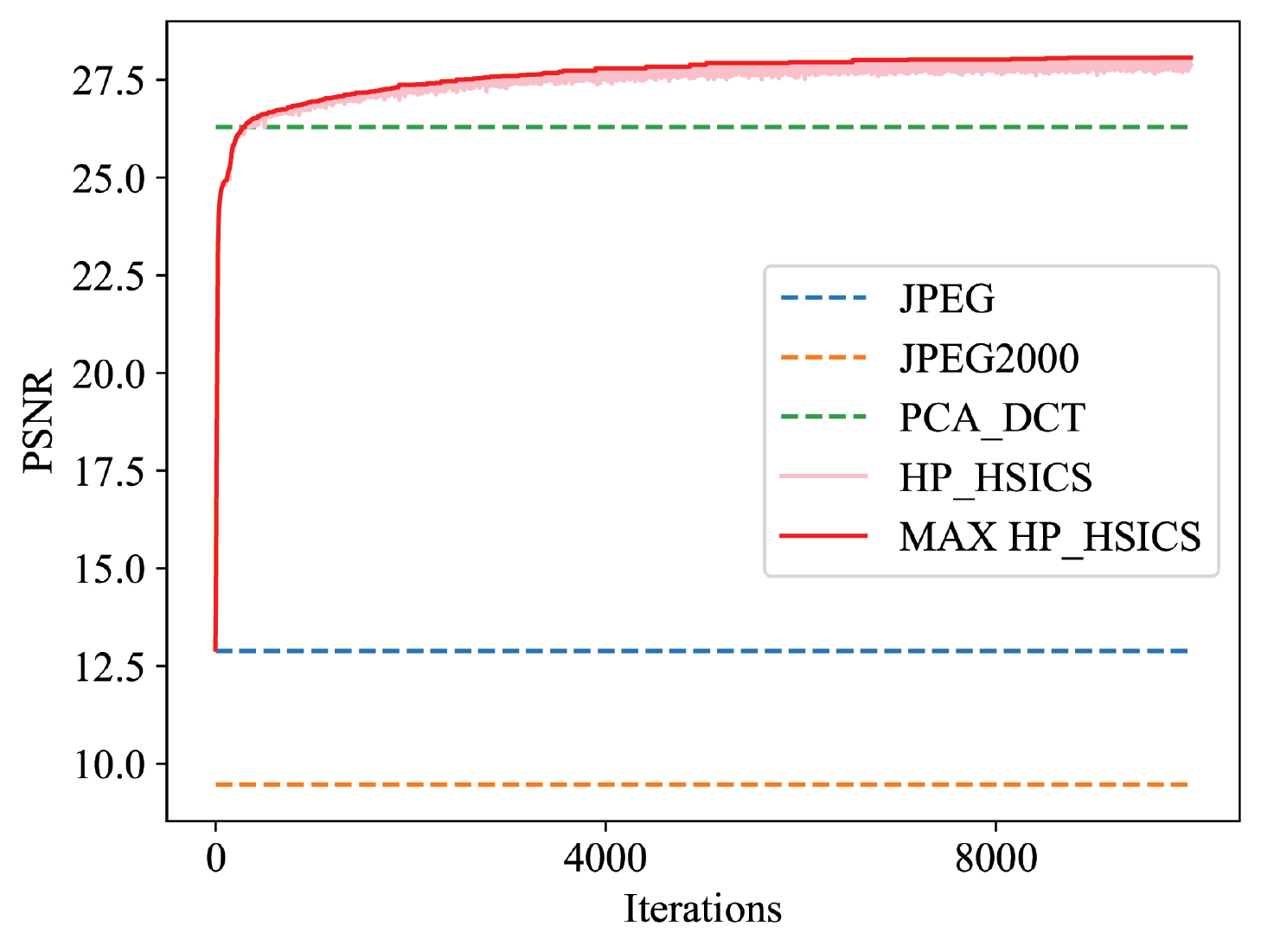}
         \caption{Model training on Cuprite dataset at 0.02 bpppb. Our method outperforms JPEG, JPEG2000, and PCA-DCT methods after some iterations and continues improving beyond that.}
         \label{fig:ip3}
     \end{subfigure}
     \hfill
     \begin{subfigure}[b]{0.46\textwidth}
         \centering
         \includegraphics[width=\textwidth]{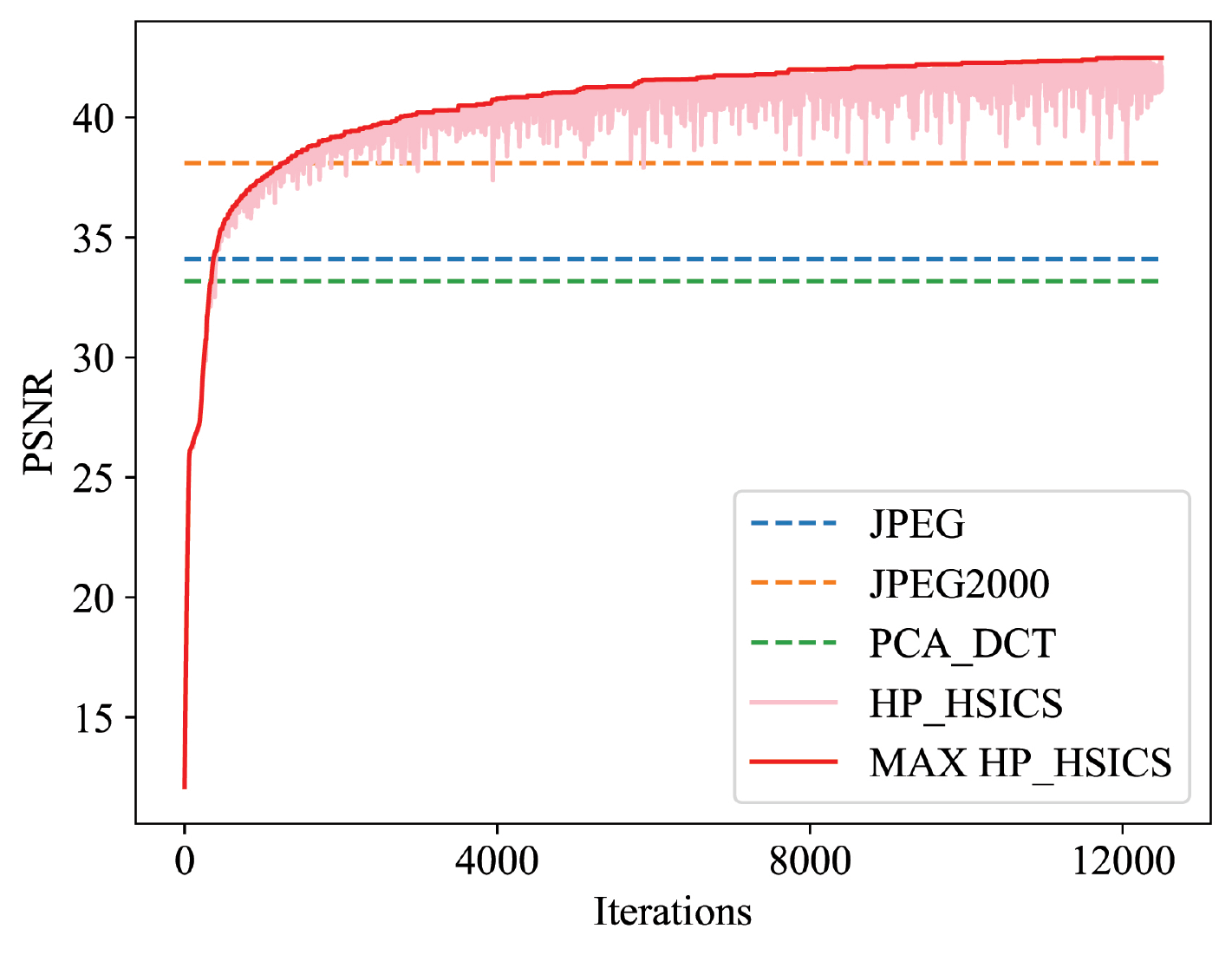}
         \caption{Model training on Indian Pines dataset at 0.2 bpppb. Our method outperforms JPEG, JPEG2000, and PCA-DCT methods after some iterations and continues improving beyond that.}
         \label{fig:ip4}
     \end{subfigure}
     \caption{Model training on different datasets}
\end{figure*}

\section{Experiments}


We compare our model against some benchmark encoders, such as JPEG \cite{good1994joint, qiao2014effective}, PCA-DCT \cite{nian2016pairwise}, and JPEG2000 \cite{du2007hyperspectral}.
JPEG 2000 uses the JPEG 2000 image compression standard and coding method to treat each band separately and encode each band independently. JPEG uses the same way to handle each band, but it uses the JPEG standard instead.
PCA-DCT employs an orthogonal transformation to reconstruct a hyperspectral image into a lower-dimensional image. The first few PCs are often used, while the other components are usually discarded after compression. The PCA-DCT-based method alters the HS image's physical composition by choosing fewer principal components. As a result, PCA-DCT was unable to achieve a higher PSNR. However, by evaluating the human vision system, PCA can still produce images with a respectable level of quality. It has been extensively utilized for HS image lossy compression uses throughout various application fields.
We choose those encoders for comparison primarily because they are all known and reliable benchmark encoders.
\par We first identify valid depth and width configurations for the MLPs representing an image to select the optimum model designs for a particular parameter budget (measured in bits per pixel per band, or bpppb). The parameter bpppb is calculated as follows:

\begin{equation}
    \label{eq:bpppb}
    bpppb = \frac{\#parameters \times (bits\ per\ parameter)}{\#pixels \times \#bands}.
\end{equation}
The optimum architecture is then chosen by doing a hyperparameter search on all viable designs and learning rates on a single image.
Each image in the dataset is used to train the final model, which is then downscaled to 16-bit precision (for model compression purposes). We observe that after training, reducing the precision of weights from 32 to 16 bits caused essentially no increase in distortion, but reducing them further to 8 bits caused a large rise in distortion, outweighing the advantage of reducing the bpppb.
\par Peak signal-to-noise ratio (PSNR) metrics are used in the experiments to compare the accuracy of our model to JPEG, JPEG2000, and PCA-DCT.
A key component of compression methods is image quality, which is determined by PSNR. The peak signal-to-noise ratio between two images is calculated using the PSNR, expressed in decibels. To contrast the level of quality of an original image compared to one that has been compressed, we utilize this ratio. As PSNR rises, the quality of the compressed or rebuilt image also rises. The efficiency of image compression is compared using the MSE and PSNR. While PSNR is a measure of the peak error, MSE is the cumulative squared error between the original and compressed image. The error decreases as the MSE value decreases. Using the following equation, we first determine the mean-squared error:
\begin{equation}
    \label{eq:mse}
    MSE = \frac{\Sigma_{M,N}(I_1(m,n)-I_2(m,n))^2}{M \times N},
\end{equation}
where M and N in the previous equation stand for the input images' respective rows and columns. The PSNR is then calculated using the following equation:
\begin{equation}
    \label{eq:psnr}
    PSNR = 10\ log_{10}\left(\frac{R^2}{MSE}\right),
\end{equation}
where R is the largest variation in the input image in the previous equation. For instance, R is 1 if the input image is of the double-precision floating-point data. R is 255, for instance, if the data is an 8-bit unsigned integer.
\par Figure \ref{fig:BP1} for the Indian Pines dataset illustrates the outcomes of this technique for various bpppb values. As can be observed, raising the bpppb level will raise the compression quality, as determined by PSNR. We did these experiments for the other datasets, Jasper Ridge in Figure \ref{fig:BP2}, Cuprite in figure \ref{fig:BP3}, and Pavia University in figure \ref{fig:BP4}, and as can be seen, the results are the same as we discussed.

\begin{figure*}
     \centering
     \begin{subfigure}[b]{0.46\textwidth}
         \centering
         \includegraphics[width=\textwidth]{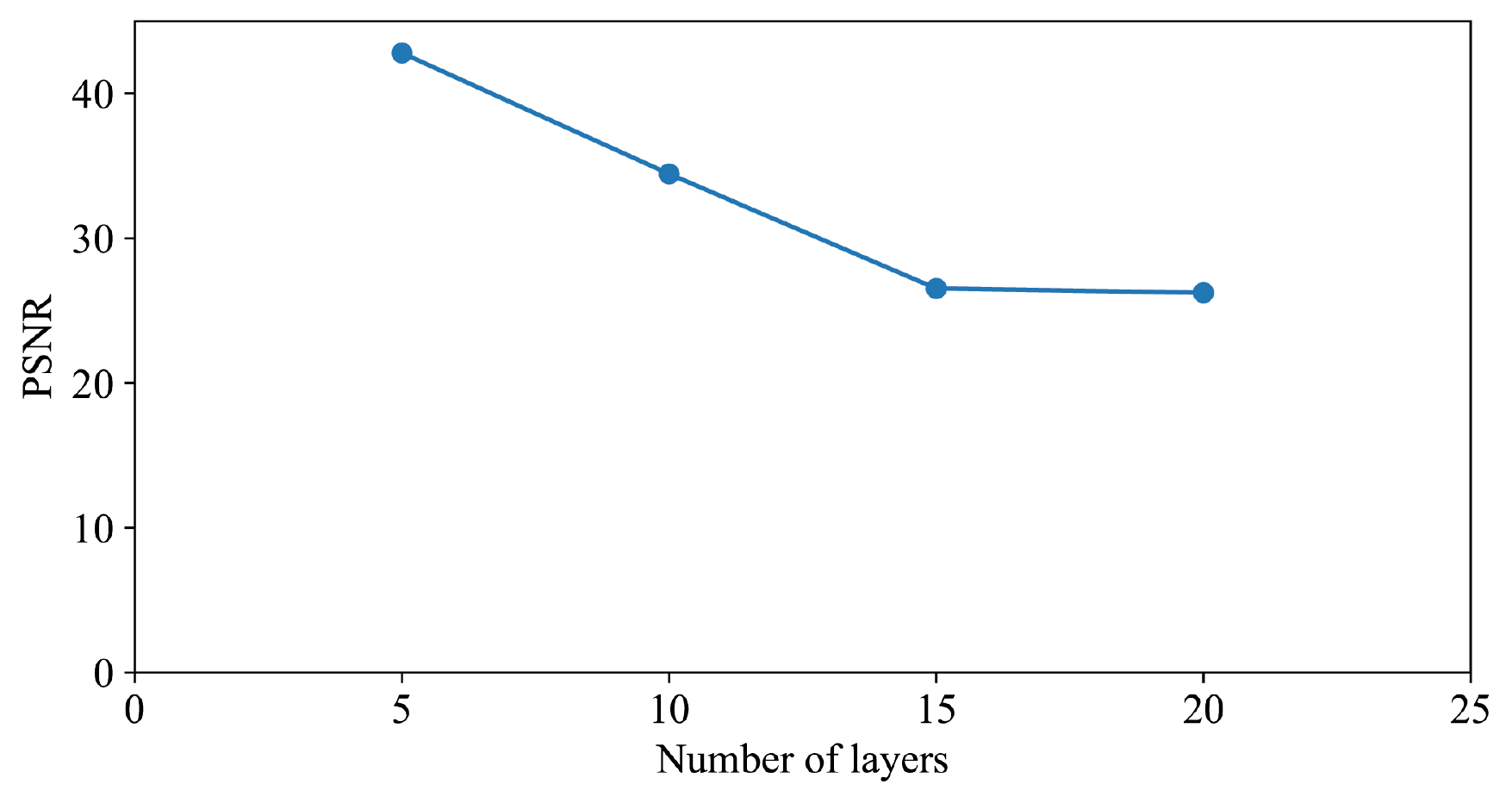}
         \caption{Indian Pines dataset}
         \label{fig:np1}
     \end{subfigure}
     \hfill
     \begin{subfigure}[b]{0.46\textwidth}
         \centering
         \includegraphics[width=\textwidth]{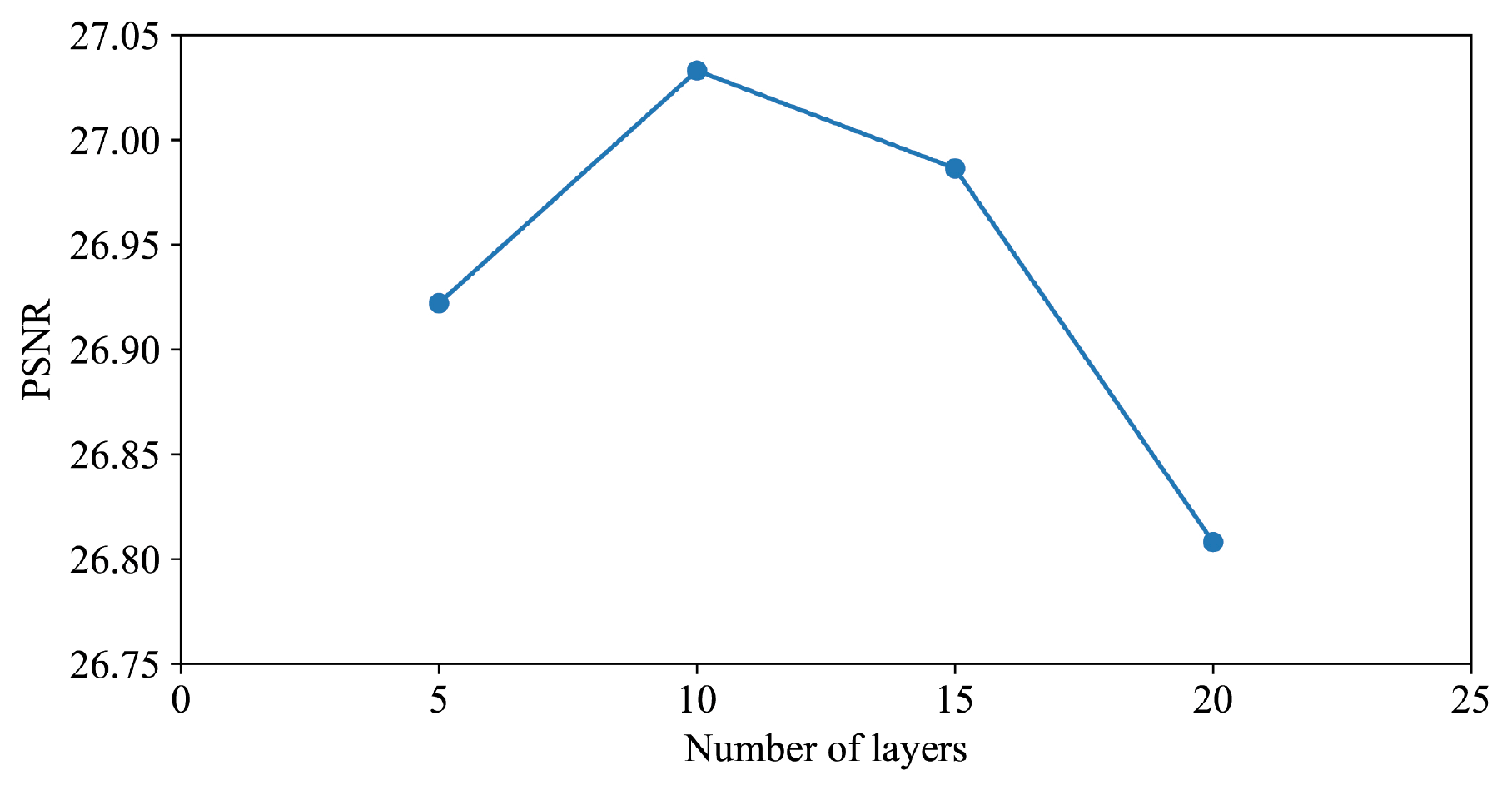}
         \caption{Cuprite dataset}
         \label{fig:np2}
     \end{subfigure}
     \hfill
     \begin{subfigure}[b]{0.46\textwidth}
         \centering
         \includegraphics[width=\textwidth]{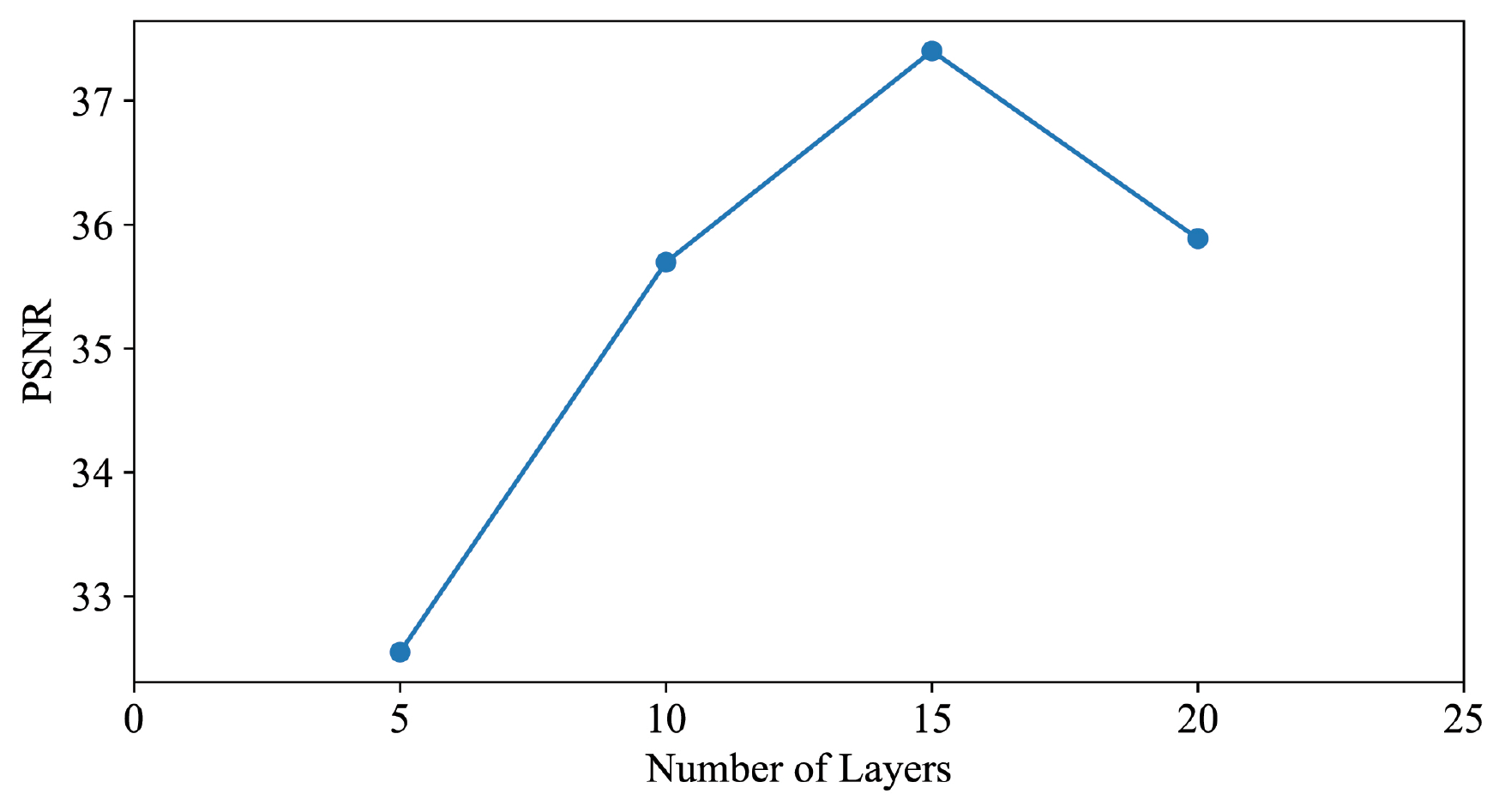}
         \caption{Jasper Ridge dataset}
         \label{fig:np3}
     \end{subfigure}
     \hfill
     \begin{subfigure}[b]{0.46\textwidth}
         \centering
         \includegraphics[width=\textwidth]{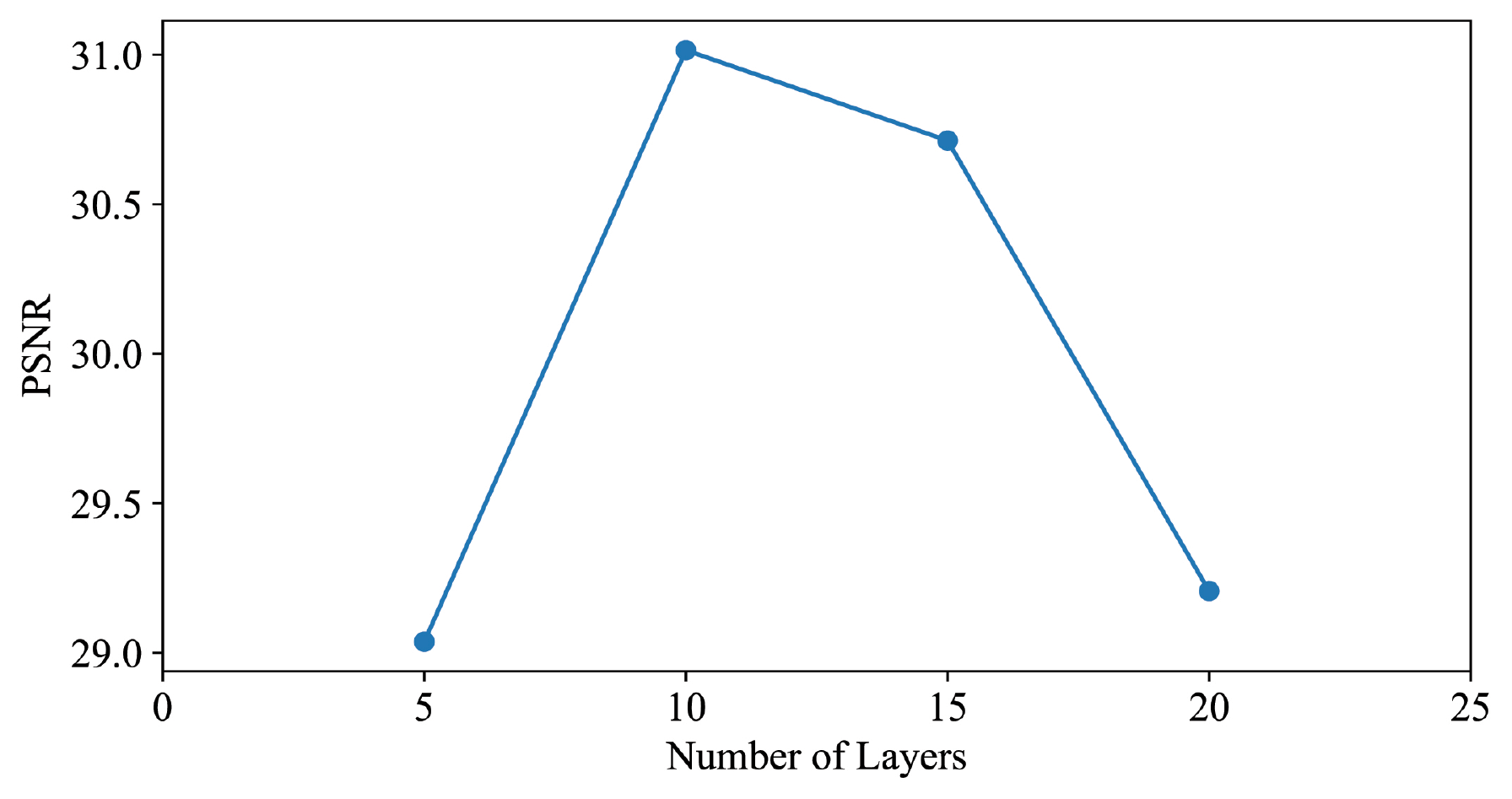}
         \caption{Pavia University dataset}
         \label{fig:np4}
     \end{subfigure}
     \caption{The performance of various valid architectures for different datasets.}
\end{figure*}




\subsection{Comparison with State-of-the-Art}
As mentioned before, we compare our method against some benchmarks. Figure \ref{fig:c1} shows this comparison for the Jasper Ridge dataset. We show the PSNRs achieved in different bpppbs for our method in two ways. We have the results with full precision (without using entropy coding), shown with HSICS; we also convert them to half-precision (using entropy coding), shown with HP\_HSICS. We then compare these results with JPEG, JPEG2000, and PCA-DCT methods. As can be seen, our method outperforms JPEG, JPEG2000, and PCA-DCT at every bpppb that we do experiments with. We also conducted another experiment on Indian Pines datasets and compared its results with JPEG2000, JPEG, and PCA-DCT. As can be seen in figure \ref{fig:c2}, our method in full precision outperforms all other methods in this comparison at every bit-rates in bpppbs, but in half-precision, it just outperforms other methods in low bit rates (lower than 0.4 bpppb). Figures \ref{fig:c3} and \ref{fig:c4} shows this comparison for Pavia University and Cuprite datasets, respectively. As can be seen, our methods outperform JPEG, JPEG2000, and PCA-DCT for the experiments on the Pavia University dataset in \ref{fig:c3}. Figure \ref{fig:c4} shows that our method outperforms JPEG, JPEG2000, and PCA-DCT, but when we want to compare the results of the half-precision method with the benchmarks, we can see that it outperforms JPEG and JPEG2000 at every bit-rates and PCA-DCT at bitrates lower than 0.02 bpppb.

\par Our approach does not need a decoder during testing, in contrast to the majority of previous neural data compression algorithms. In fact, the decoder model is enormous even though the latent code representing the compressed image in such systems is small. As a result, the decoding device also needs a lot of memory. In our situation, the decoder side memory requirements are orders of magnitude fewer because we only need the weights of a very small MLP.
We show an example of the overfitting procedure in Figures \ref{fig:ip1}, \ref{fig:ip2}, \ref{fig:ip3}, and \ref{fig:ip4}. This experiment has been done on Jasper Ridge at 0.15 bpppb, Pavia University at 0.025 bpppb, Cuprite at 0.02 bpppb, and Indian Pines at 0.2 bpppb. We compare our model with JPEG, JPEG2000, and PCA-DCT methods. As can be seen, our method outperforms JPEG, JPEG2000, and PCA-DCT methods after some iterations and continues improving beyond that.
While the optimization can be noisy, we simply save the model with the best PSNR. We just store the model with the best PSNR, given the fact that optimization can be noisy.
\subsection{Experimental Details}
In Figures \ref{fig:np1}, \ref{fig:np2}, \ref{fig:np3}, and \ref{fig:np4}, we show the performance of various valid architectures of size 0.2 bpppb for Indian Pines dataset, 0.01 bpppb for Cuprite dataset, 0.1 bpppb for Jasper Ridge dataset, and 0.03 bpppb for Pavia University dataset, respectively. As can be observed, the design choice affects the compression quality, with different optimal architectures for various bpppb values. In the following, we will explain the experimental detail of this work.
\par
\begin{figure}
\begin{center}
    \includegraphics[width=0.4\linewidth]{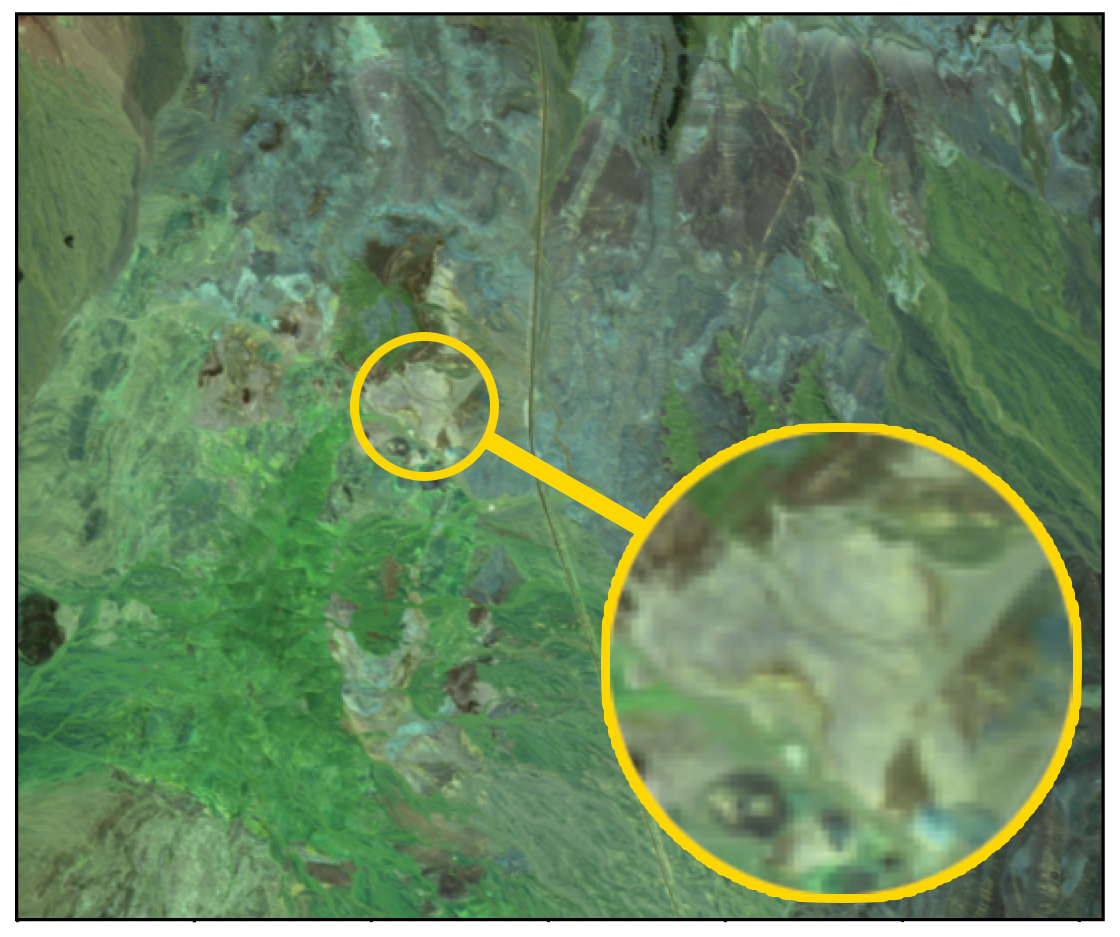}
    \includegraphics[width=0.4\linewidth]{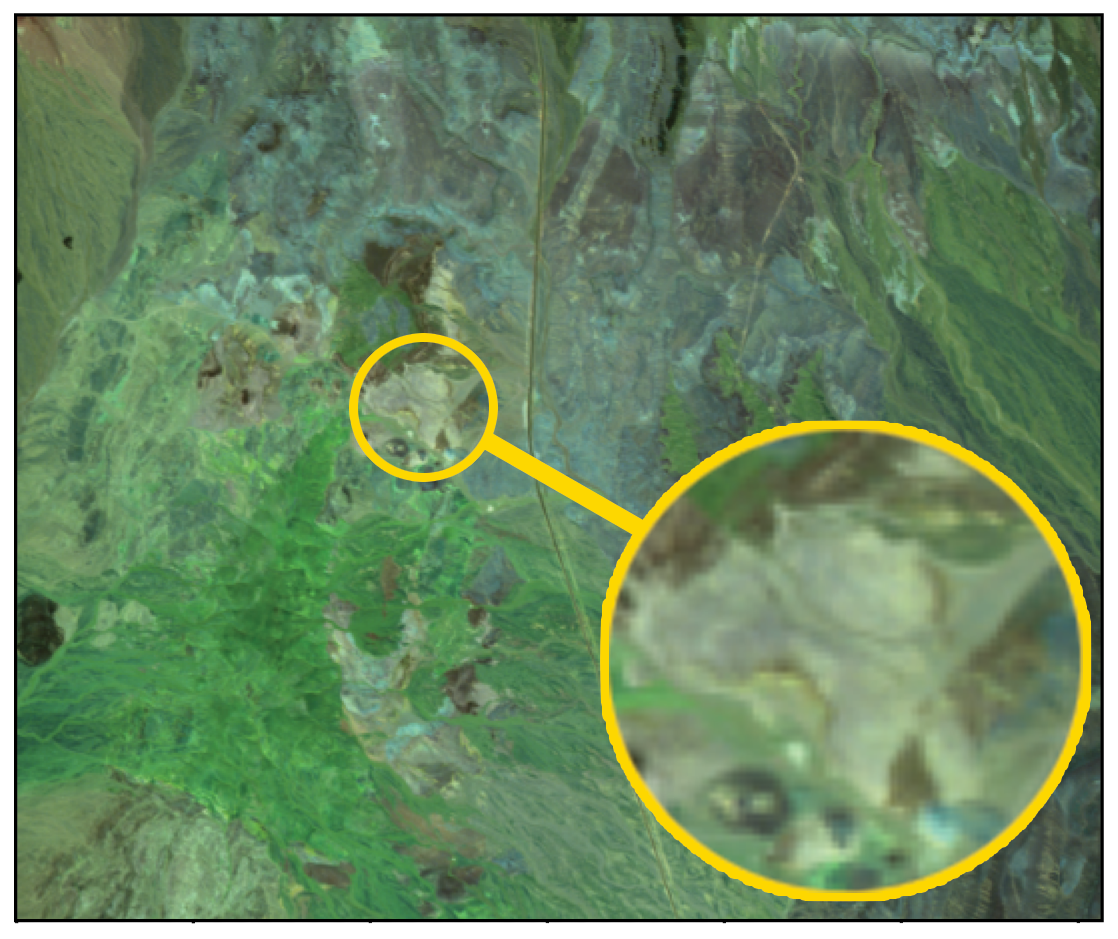}
\end{center}
   \caption{Cuprite original image in the left and the reconstructed image in the right.}
\label{fig:m7}
\end{figure}

\begin{figure}
\begin{center}
    \includegraphics[width=0.4\linewidth]{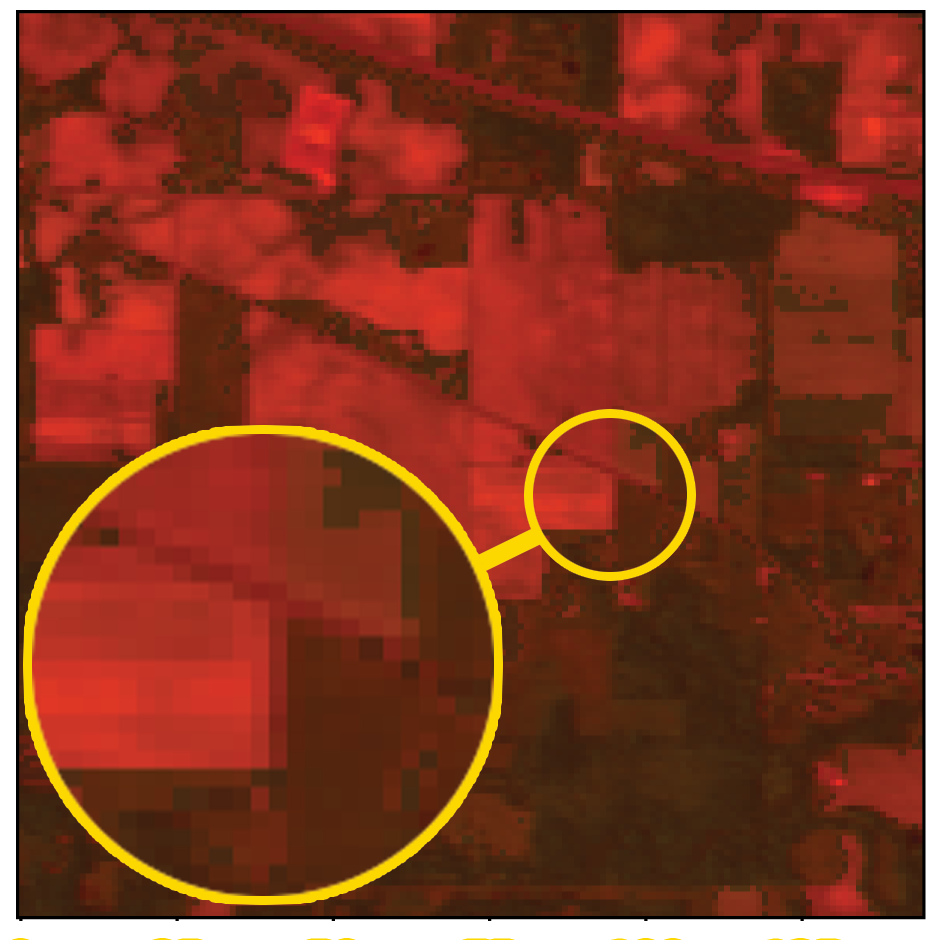}
    \includegraphics[width=0.4\linewidth]{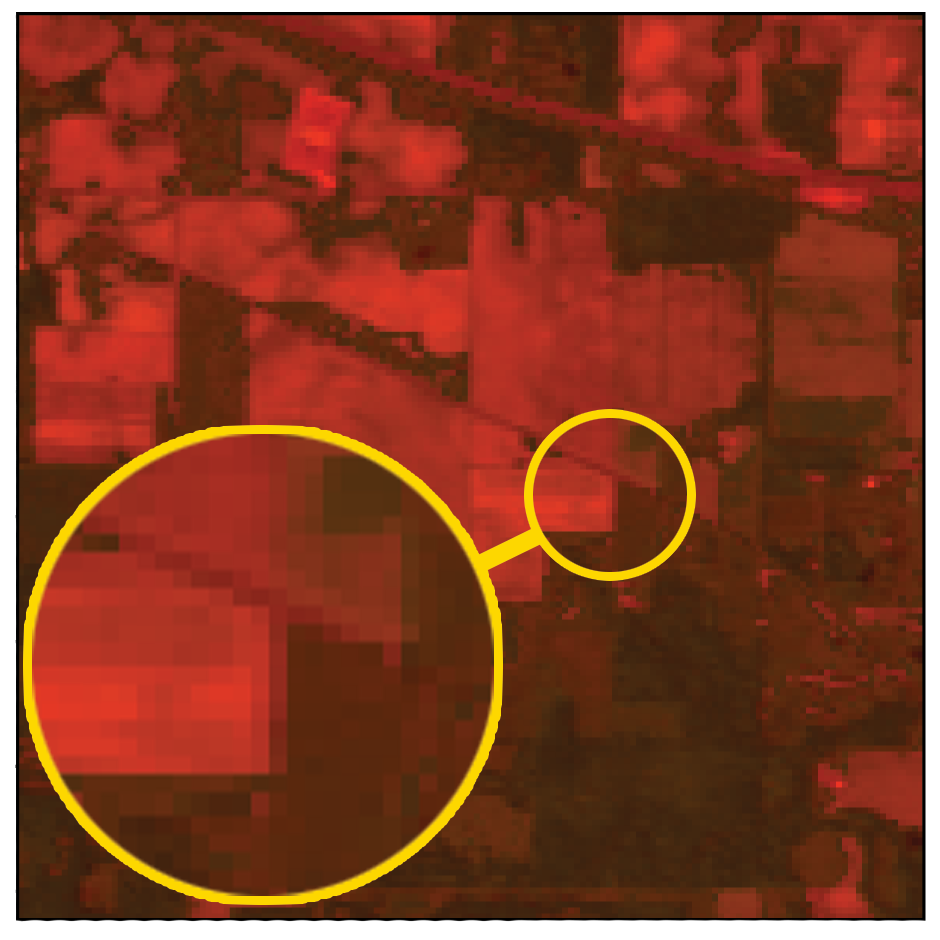}
\end{center}
   \caption{Indian Pines original image in the left and the reconstructed image in the right.}
\label{fig:m8}
\end{figure}

\begin{figure}
\begin{center}
    \includegraphics[width=0.4\linewidth]{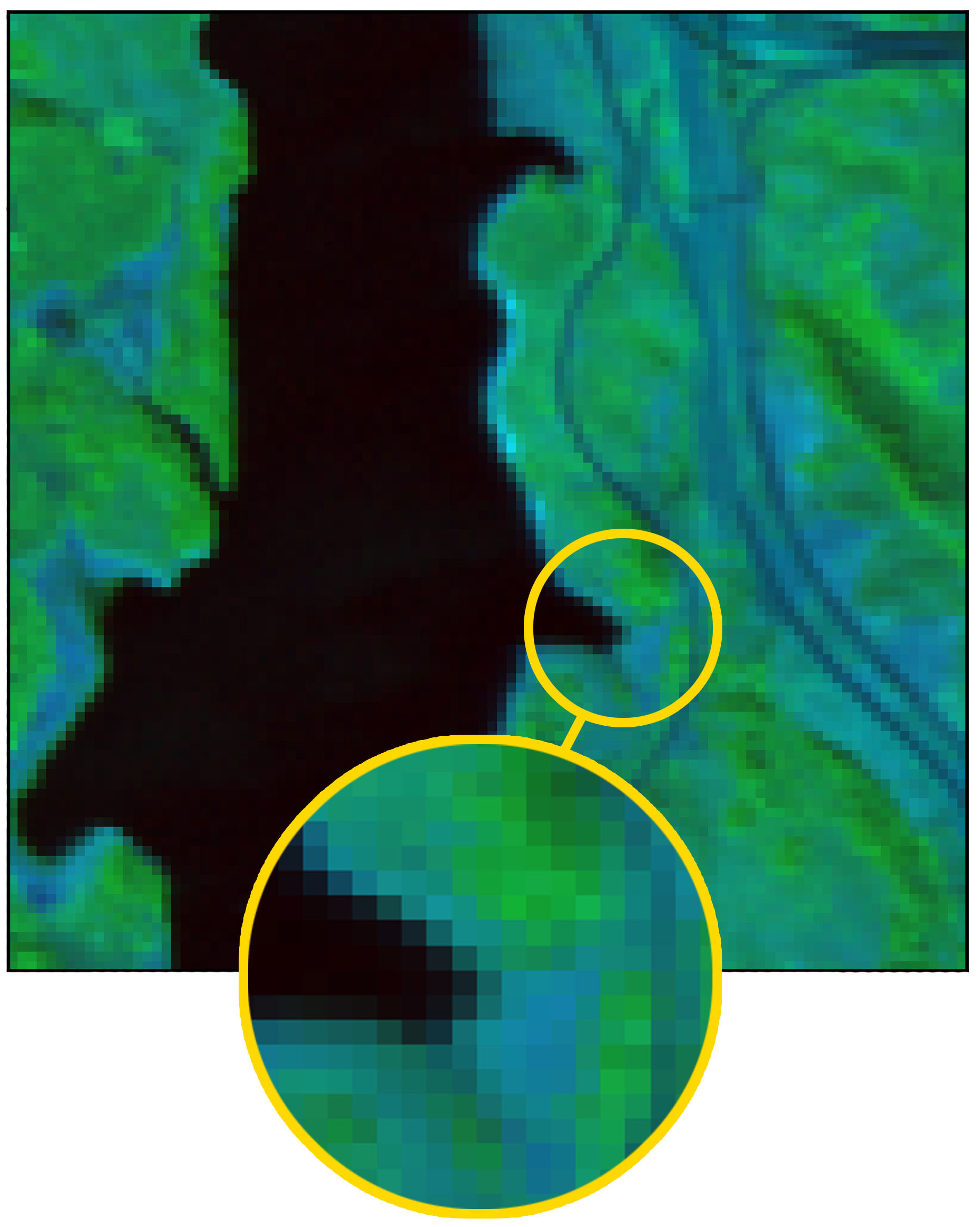}
    \includegraphics[width=0.4\linewidth]{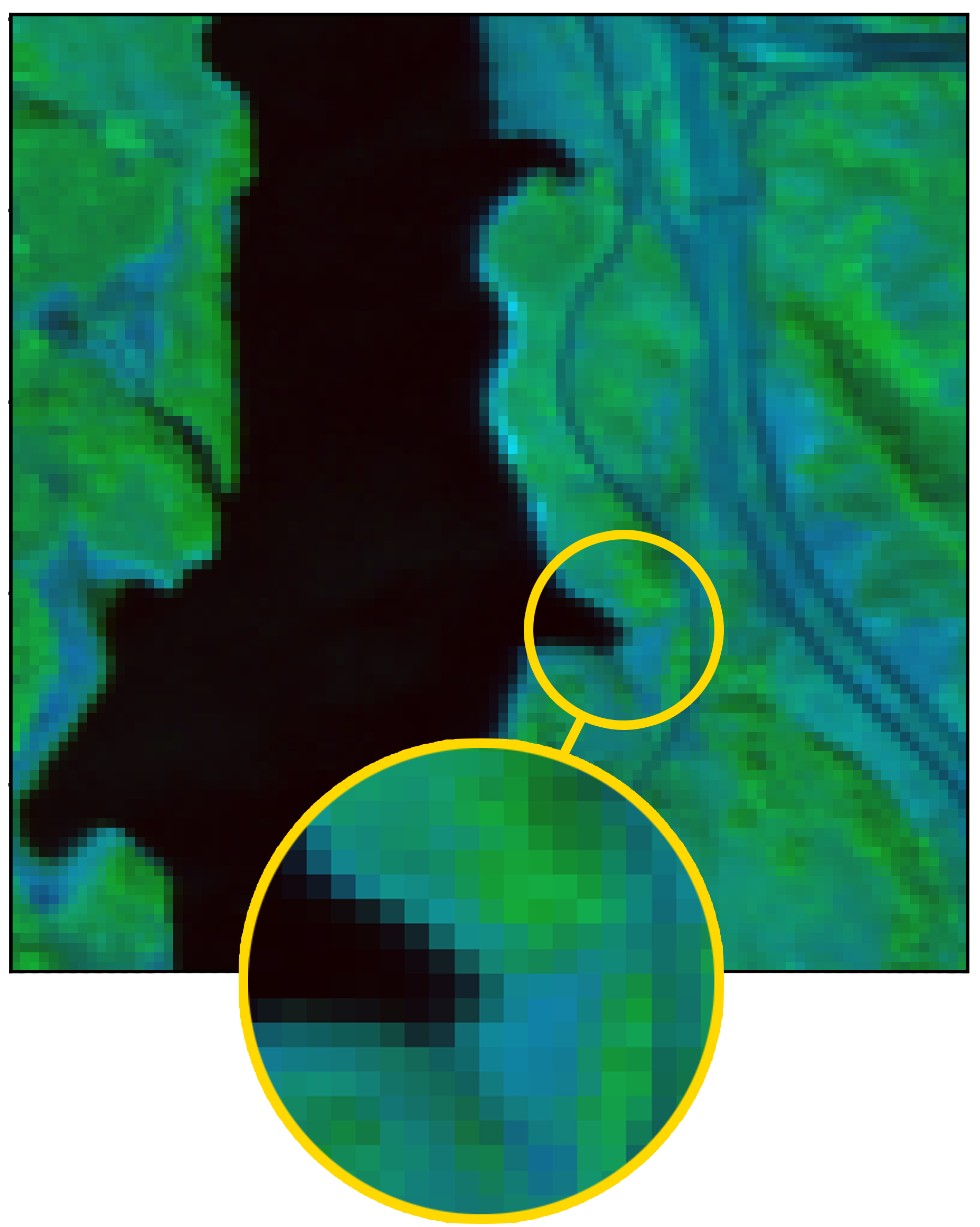}
\end{center}
   \caption{Jasper Ridge original image in the left and the reconstructed image in the right.}
\label{fig:m9}
\end{figure}

\begin{figure}
\begin{center}
    \includegraphics[width=0.4\linewidth]{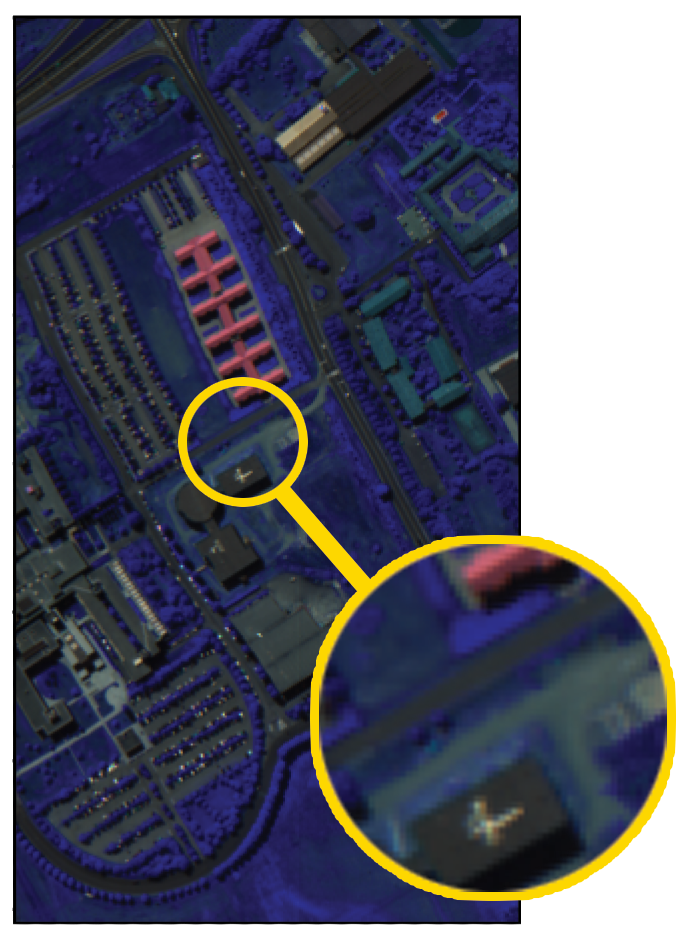}
    \includegraphics[width=0.4\linewidth]{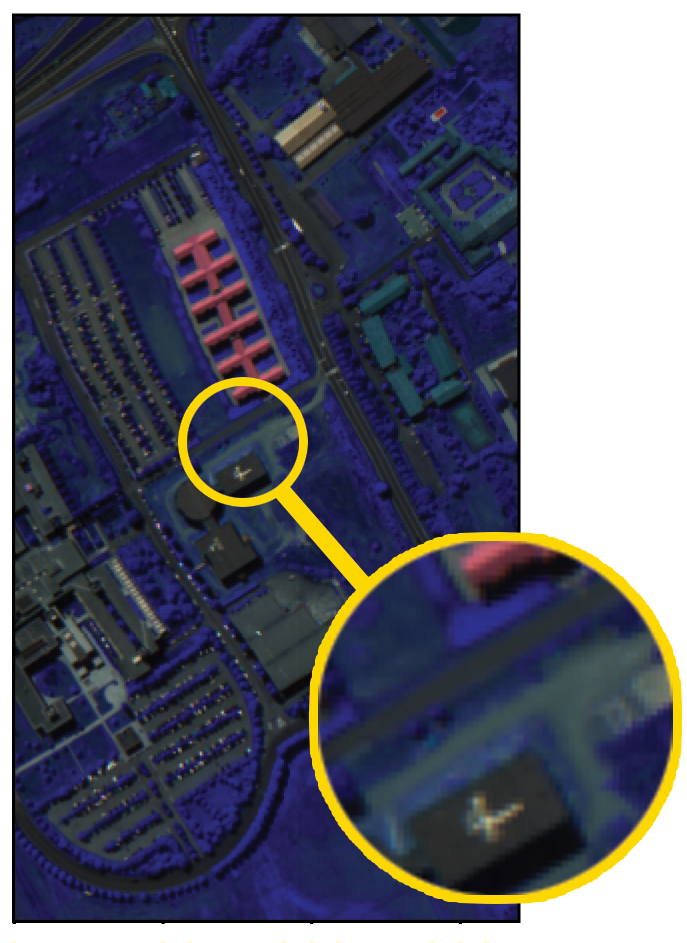}
\end{center}
   \caption{Pavia University original image in the left and the reconstructed image in the right.}
\label{fig:m10}
\end{figure}

 Adam was used to training all models. We employed MLPs with output dimensions related to the amount of dataset image bands and two input dimensions (equivalent to (x, y) coordinates). Except for the final layer, we used sine non-linearities and the initialization method given in \cite{sitzmann2020implicit}. A learning rate of 2e-4 was employed. In figures \ref{fig:m7}, \ref{fig:m8}, \ref{fig:m9}, \ref{fig:m10}, we include qualitative results comparing the compression artifacts from our proposed method and the original image on the Cuprite, Indian Pines, Jasper Ridge, and Pavia University datasets, respectively.

\section{Conclusion}
In this work, we employ implicit neural representation to compress hyperspectral images by using neural networks to map pixel locations to band values and store the weights of the generated networks. We then assess the MLP at each pixel position to decode the image. Through experiments, we demonstrated that this method can outperform JPEG, JPEG2000, and PCA-DCT at low bit rates. For future work, we want to use meta-learned base networks, a new method for neural image compression, and then add PCA to that method to improve the results.
{\small
\bibliographystyle{ieee_fullname}
\bibliography{egbib}
}

\end{document}